\documentclass[sts]{imsart}

\RequirePackage{amsthm,amsmath,amsfonts,amssymb}
\RequirePackage[numbers,sort&compress]{natbib}
\RequirePackage[colorlinks,citecolor=blue,urlcolor=black]{hyperref}
\RequirePackage{graphicx}

\usepackage{graphicx}

\usepackage{multirow}
\usepackage{setspace}

\usepackage{etoolbox}

\setcitestyle{numbers}
\setcitestyle{square}
\usepackage{etoolbox}

\usepackage{hyperref}

\usepackage{algorithm,algorithmic}
\usepackage{setspace}
\usepackage{tikz}

\startlocaldefs
\theoremstyle{plain}

\newtheorem{theorem}{Theorem}[section]

\theoremstyle{remark}
\newtheorem{definition}[theorem]{Definition}

\newcommand{\argmax}{\operatorname*{arg \ max}}
\newcommand{\argmin}{\operatorname*{arg \ min}}

\newcommand{\bo}{\boldsymbol}
	
\newcommand{\indep}{\perp \!\!\! \perp}

\newcommand{\iid}{\overset{\mathrm{iid}}{\sim}}

\AtBeginEnvironment{algorithm}{\setstretch{1.35}}

\usepackage{amsmath}

\endlocaldefs

\begin{document}

\begin{frontmatter}
\title{Bayesian Transfer Learning}
\runtitle{Bayesian Transfer Learning}
\runauthor{P. Suder, J. Xu, and D. Dunson}

\begin{aug}
\author[A]{\fnms{Piotr M.}~\snm{Suder}\ead[label=e1]{piotr.suder@duke.edu}},
\author[B]{\fnms{Jason}~\snm{Xu,}\ead[label=e2]{jason.q.xu@duke.edu}}
\and
\author[C]{\fnms{David}~\snm{B. Dunson}\ead[label=e3]{dunson@duke.edu}}


\address[A]{Piotr M. Suder: PhD Student, Department of Statistical Science, Duke University \printead[presep={\ }]{e1}.}

\address[B]{Jason Xu: Assistant Professor, Department of Statistical Science, Duke
University \printead[presep={\ }]{e2}.}

\address[C]{David B.
Dunson: Arts and Sciences Distinguished Professor,
Departments of Statistical Science and Mathematics, Duke
University \printead[presep={\ }]{e3}.}

\end{aug}

\begin{abstract}
Transfer learning is a burgeoning concept in statistical machine learning that seeks to improve inference and/or predictive accuracy on a domain of interest by leveraging data from related domains. While the term "transfer learning" has garnered much recent interest, its foundational principles have existed for years under various guises. Prior literature reviews in computer science and electrical engineering have sought to bring these ideas into focus, primarily surveying general methodologies and works from these disciplines. This article highlights Bayesian approaches to transfer learning, which have received relatively limited attention despite their innate compatibility with the notion of drawing upon prior knowledge to guide new learning tasks. Our survey encompasses a wide range of Bayesian transfer learning frameworks applicable to a variety of practical settings. We discuss how these methods address the problem of finding the optimal information to transfer between domains, which is a central question in transfer learning. {We illustrate the utility of Bayesian transfer learning methods via a simulation study where we compare performance against frequentist competitors.} 
\end{abstract}


\end{frontmatter}

\section{Introduction}

\textit{Transfer learning}---applying knowledge gained from training on previous tasks and domains toward new tasks---is a burgeoning concept in statistics and machine learning. This natural idea mimics some of the mechanisms of human intelligence where past experience, skills and knowledge are often utilized in learning new topics. It is appealing to apply the same paradigm in developing machine intelligence to extract knowledge from the rapidly growing body of datasets available to scientists which are often related to each other in various ways. If the domains between which the transfer of information occurs are sufficiently related,  transfer learning can substantially improve the performance of the target model. This is particularly useful when we have a small target dataset we want to study which does not contain enough datapoints to extract precise inferences or predictions, but have access to a large, related dataset. 

For instance, suppose we want to study brain connectomes of Alzheimer's patients or genomes of people suffering from a rare type of cancer. We may utilize large datasets of brain connectomes or genomes collected from healthy individuals such as the ones provided by the UK Biobank to improve the models fitted to the target data. These related sources may aid in the extraction of, say, a low dimensional latent representation of the complex data we seek to study, which can be useful toward dimensionality reduction in the target domain. 

Although the term \textit{transfer learning} has seen increasing popularity in recent years, some of the ideas undergirding it have been around for much longer, and have appeared under various names. Several recent literature reviews 
aim to help researchers organize and classify these ideas systematically. To name a few, \cite{survey_transfer_learning}, and more recently \cite{survey_tl_2020} and \cite{ieee_survey_tl}, provide general overviews on transfer learning methodology, largely from the computer science and electrical engineering literature. \cite{deep_tl_survey} focuses on transfer learning in deep neural networks, an appealing use-case due to the data-hungry nature of deep learning models together with the availability of large datasets for training source models. Areas where deep learning is commonly applied such as computer vision often leverage public datasets such as ImageNet \cite{imagenet} or Open Images V4 \cite{open_images_v4}, with millions of datapoints available for training. 
Meanwhile, \cite{negative_transfer_survey} focuses on the phenomenon of negative transfer, where the source domains 
are too different from the target domain, so that applying transfer learning \textit{worsens} the performance of the target learner. The existence of the negative transfer phenomenon illustrates the importance of choosing an appropriate amount of information to be transferred (the "strength" of transfer) between domains, which remains one of the key challenges in transfer learning and will be one of the focal topics in this survey. 

With the exception of \cite{xuan2021bayesian}, none of these literature reviews substantially focus on Bayesian views. While the work of Xuan \textit{et. al} \cite{xuan2021bayesian} explicitly overviews Bayesian transfer learning, its scope is limited to probabilistic graphical models. 
One can argue that the Bayesian paradigm provides a natural framework for how to incorporate prior information from previous datasets within current inferences, and hence provides a canonical umbrella of approaches for transfer learning.
In this paper we provide an overview of some highlights of the Bayesian transfer learning literature. Our focus is on describing how different classical Bayesian approaches can be either directly applied or easily adapted to transfer learning problems.
In doing so, we contribute various ideas toward answering a central question of transfer learning: how do we determine and enforce optimal information transfer between domains
utilizing various Bayesian modeling approaches?
Our aim is to contribute a broad view of Bayesian transfer learning, while presenting approaches that help surmount the problem of negative transfer.

The rest of the paper is organized as follows. In the following section, we give formal definitions of transfer learning and related areas, and discuss alternative names for related ideas in the literature. In Section \ref{methods_sec} we provide an overview of general Bayesian approaches to transfer learning with specific examples and some applications. In Sections \ref{interp_intra_sec} and \ref{data_labels_sec} we provide a brief taxonomy of transfer learning and point out several areas where some specific Bayesian approaches introduced here can be particularly useful. Finally, in Section \ref{simulation_study} we present a simulation study comparing one of the Bayesian methods introduced here with frequentist transfer learning competitors. We conclude with a discussion in Section \ref{discussion}.

\section{Definition and related areas}\label{def_section}

While approaches for transferring information across statistical tasks have a rich history, use of the ``transfer learning'' terminology is relatively recent. Perhaps as a result, there is not yet a standard technical definition of what qualifies as transfer learning.
While some authors adopt a narrow definition of transferring parameters between models \cite{metalearning_survey}, others welcome broader, more general definitions \cite{survey_transfer_learning}, \cite{ieee_survey_tl}, \cite{deep_tl_survey}. In this section we provide one definition of transfer learning to fix ideas for the rest of the article. We then discuss closely related areas. Here by \textit{domain} we denote the two-element set of the form $\mathcal{D} = \{\mathcal{X}, P\}$, where $\mathcal{X}$ is the \textit{feature space} and $P$ is the marginal probability distribution of the observations $X \in \mathcal{X}$ collected in a dataset associated with $\mathcal{D}$. Given a domain $\mathcal{D}$ and its associated \textit{label space} $\mathcal{Y}$, \cite{survey_transfer_learning} define a \textit{task} on $\mathcal{D}$ as the set $\mathcal{T} = \{\mathcal{Y}, f(\cdot)\}$, where $f$ is a function given by $f = \{(x,y) \mid x \in \mathcal{X}, y \in \mathcal{Y} \}$. 
In this framework, $f$ is the ground truth, the optimal solution to the task which is not directly observed but whose approximation can be learned from the observed data.

\begin{definition}[\textbf{Transfer Learning}]\label{def_tl}
    Consider the \textit{source domains} $\mathcal{D}_{1}, \mathcal{D}_{2}, \dots, \mathcal{D}_{K}$ with respective associated \textit{source tasks} $\mathcal{T}_{1}, \mathcal{T}_{2}, \dots, \mathcal{T}_{K}$, as well as the \textit{target domain} $\mathcal{D}_0$ with the associated \textit{target task} $\mathcal{T}_0 = \{\mathcal{Y}_0, f_0\}$, where an approximation to $f_0$ can be learned based on the available data $(X_0, Y_0)$ with $X_0 \in \mathcal{X}_0, Y_0 \in \mathcal{Y}_0$. Suppose that $\mathcal{D}_{k} \neq \mathcal{D}_{0}$ or $\mathcal{T}_{k} \neq \mathcal{T}_{0}$ for any $k = 1,\dots,K$. \textbf{Transfer learning} refers to algorithms which aim at improving the approximation of $f_0$ by incorporating the knowledge from $\mathcal{D}_{1}, \mathcal{D}_{2}, \dots, \mathcal{D}_{K}$ and $\mathcal{T}_{1}, \mathcal{T}_{2}, \dots, \mathcal{T}_{K}$.

\end{definition}
\noindent In this setting, by \textit{knowledge} we mean either: (i) the raw data sampled from the source domains, possibly equipped with labels from the source tasks, (ii) learners pre-trained on the data from source domains and tasks, or (iii) oracle models which have complete and true information on the source domains and the source tasks. 
Among these, cases (i) and (ii) are the most commonly encountered ones in practice.

Our definition follows and generalizes the conventions used by \cite{survey_transfer_learning}. 
Note that in the above definition, the source and target domains are not necessarily different, encompassing cases with a common domain but different tasks. Furthermore, when two domains are different, their feature spaces need not differ. 
In the deep learning literature the target task is sometimes referred to as the \textit{downstream task} \cite{pretrain_loss_deep}.

By allowing the label space to be (a) discrete, (b) one-dimensional, (c) multidimensional, (d) defining only a partition of a dataset associated with $\mathcal{D}$ without giving a specific meaning to how the labels are used for that purpose, this definition comprises, respectively, (a) classification, (b) univariate regression, (c) multivariate regression and dimensionality reduction, and (d) clustering tasks. Finally, allowing the values of $f$ to be probability distributions naturally lends itself to  Bayesian posterior learning.

\subsection{Related fields}
    Another closely related problem which follows this paradigm  is \textit{multitask learning}. Just like transfer learning, its goal is to improve learning on a particular domain based on information from related domains and tasks, but it differs in seeking to simultaneously learn each task jointly on all the domains considered. This may improve the performance across tasks by borrowing information across related tasks and domains, in contrast to using a set of tasks and domains only as means to the end of improving performance on a single \textit{target task} \cite{survey_transfer_learning}. While some authors regard multitask learning and transfer learning as separate disciplines \cite{survey_transfer_learning}, \cite{yang_zhang_dai_pan_2020}, \cite{Wang_Pineau_2015}, others either consider them as the same field \cite{xuan2021bayesian}, or draw a distinction between the two according to different criteria, as in \cite{kernel_bayes_tl}. 
    Often a multitask learning method can be easily adapted to transfer learning \cite{survey_transfer_learning}, \cite{yang_zhang_dai_pan_2020}. As we will see in the following sections, the Bayesian framework elegantly reconciles these notions in many cases. 

    \textit{Continual} or \textit{lifelong learning} \cite{continual_learning_1}, \cite{continual_learning_2}, \cite{continual_learning_1}, \cite{continual_learning_3}, \cite{continual_learning_4} is a popular concept in machine learning,  which combines aspects of transfer and multitask learning. In this setting an agent faces a sequence of domain-task pairs over time, with the goal being to  utilize previously encountered tasks to learn each new task in a more effective way while maintaining the ability to solve the previous tasks \cite{reinforced_cont_learning}. Continual learning attempts to provide a remedy to the phenomenon of catastrophic forgetting \cite{cat_forgetting} in transfer learning where the model performs worse on the source tasks after being adjusted to the target task; this commonly occurs in deep learning models \cite{deep_forgetting}, \cite{  gan_forgetting}.
    This forgetting can be especially problematic when the number of encountered tasks and model parameters becomes large and it becomes difficult to store the previously encountered datasets and models trained on them.

There is additionally a Bayesian literature on \textit{metalearning}, or \textit{learning to learn}  \cite{bayesian_metalearning}, \cite{amortized_meta_few_shot}, 
    \cite{ravi2018amortized},
    \cite{continual_metalearning}. Vanschoren \cite{Vanschoren2019} defines metalearning as methods aiming to improve the ``configuration'' (e.g. model hyperparameters, network architecture in case of deep learning methods, etc.) of the model for the target task by training on \textit{metadata}. 
    Here metadata refers to information obtained from models trained individually {with} different configurations; for example, one may vary different aspects of the model and measure its performance via cross validation. While some researchers consider metalearning as distinct from transfer learning \cite{metalearning_survey}, following Definition 1 we consider it to be a special case of transfer learning. In this case the information from the source domains is utilized by training models with various configurations on these domains and then using the metadata generated from them in improving the model for the target task.  
    

    \textit{Domain adaptation} \cite{bayes_cov_shift}, \cite{cancer_rna}, \cite{germain2013}, \cite{bayes_language_da} is another popular term, which is sometimes used interchangeably with transfer learning as in \cite{domain_adapt_tl}. However, since knowledge can also be transferred between different tasks on the same domain, we view it as a particular case of transfer learning. 
    Additional terminology for concepts closely related to transfer learning includes cooperative learning \cite{multiple_learners_graph}, \cite{tibshiriani_cooperative_learn}, knowledge consolidation,
context-sensitive learning, knowledge-based inductive bias,
incremental, and cumulative learning \cite{survey_transfer_learning}. 





\section{Bayesian approaches to transfer learning}\label{methods_sec}
Two fundamental questions which need to be addressed are: (i) {\em how} information should be transferred, and (ii) {\em which} information should be transferred. There are various approaches to answering these questions and they are often related to the models used for solving the source and target tasks. Determining appropriate information transfer between domains is critical, since transferring inappropriate information can result in large bias and suboptimal performance. In extreme cases, one obtains \textit{negative transfer} \cite{negative_transfer_survey}, which corresponds to the case in which transferring information decreases performance.

Some of the existing approaches rely on expert knowledge about the domains considered and their relationships, some introduce statistical measures of similarity between domains, while others rely on more flexible model-based or validation-based approaches to the optimal choice of parameters controlling information transfer. In this section we discuss different ideas based on Bayesian methodology which can be used to tackle questions (i)-(ii).

\subsection{Shared parameters}\label{shared_params_section}
One of the most prevalent approaches is to use common parameters in the source and target domains. For exposition, throughout this subsection we assume only one source dataset $X_S$ and one target dataset $X_T$. For convenience of notation, by $X_d$ we denote both the datapoints and their associated labels (when applicable) for domain $d \in \{S,T\}$. We parameterize the data likelihood for both source and target domains as $p(X_S \mid \theta_{C}, \theta_{S})$ and $p(X_T \mid \theta_{C}, \theta_{T})$, respectively, where $\theta_C$ is the common vector of parameters, shared by both source and target data, while $\theta_S$ and $\theta_T$ are vectors of parameters unique to the datasets. Let $\pi(\theta_C)$, $\pi(\theta_S)$, $\pi(\theta_T)$ be the prior distributions for, respectively, $\theta_C$, $\theta_S$, $\theta_T$. 

A simple Bayesian transfer learning approach would compute the posterior for $\theta_C$ based on the prior $\pi(\theta_C)$ and the source data $X_S$ via 
\begin{eqnarray}
p(\theta_C \mid X_S) &\propto & p(X_S \mid \theta_C)\pi(\theta_C) \nonumber \\ 
&=& \left(\int p(X_S \mid \theta_C, \theta_S)\pi(\theta_S) d\theta_S \right) \pi(\theta_C) ,\label{eq:posterior_source}   
\end{eqnarray}
and then use $\pi^*(\theta_C, \theta_T) \propto 
p(\theta_C \mid X_S) \pi(\theta_T)$ as the prior for $(\theta_C, \theta_T)$ in the analysis of $X_T$ to obtain the posterior 
\begin{equation*}
p^*(\theta_C, \theta_T \mid X_T) \propto p(X_T \mid \theta_C, \theta_T)\pi^*(\theta_C, \theta_T).
\end{equation*}
It is straightforward to see that if $X_T \indep X_S \mid (\theta_C, \theta_T)$ and $\theta_S \indep \theta_T$ a priori, then this is equivalent to obtaining a posterior for $(\theta_C, \theta_T)$ based on the data $(X_T, X_S)$ with the prior $\pi(\theta_C)\pi(\theta_T)$ on $(\theta_C, \theta_T)$, i.e.
\begin{equation}\label{eq:bayes_equiv}
    p^*(\theta_C, \theta_T \mid X_T) = p(\theta_C, \theta_T \mid X_T, X_S),
\end{equation}
where 
$$p(\theta_C, \theta_T \mid X_T, X_S) \propto p(X_T, X_S \mid \theta_C, \theta_T)\pi(\theta_C)\pi(\theta_T).$$
Hence, this approach is equivalent to giving equal weights to the source and target data in computing the posterior of the shared parameters. This is an appropriate approach when the model is correctly specified and the true parameters $\theta_C$ are indeed exactly the same in the source and target populations. 

However, in practice, it is likely that the assumption of exactly equivalent values of $\theta_C$ is an oversimplification. As the true values of $\theta_C$ vary more widely between the source and target domains, the above approach can have suboptimal performance, particularly when the source data sample size is larger than that of the target, which is often the case. A simple and commonly used heuristic solution is to specify the prior for $\theta_C$ in the target posterior as a variance-inflated version of the posterior 
$p(\theta_C \mid X_S)$ 
from the source data analysis.

Shwartz-Ziv \textit{et al.} \cite{pretrain_loss_deep} apply a related approach to Bayesian deep neural networks (DNNs). First a Gaussian approximation to the posterior for the DNN fitted to the source data is obtained.  The authors assume the ``feature extractor'' layers of the DNN are common to the source and target DNN. The variance of the Gaussian approximation to the source posterior for the weights in these layers is scaled up by a constant factor and then used as a prior for the feature extractor component in the target data DNN. 
The remaining weights characterizing the ``head'' of the DNN are given an isotropic Gaussian prior. 
To learn an appropriate amount of information sharing between the source and target domain, the scaling factor is chosen on held-out validation data from the target training dataset.



An alternative approach to controlling the influence of the source data on the target domain posterior distribution is the \textit{power prior} \cite{power_prior_1}, \cite{partial_borrowing_power_prior}, \cite{optimality_power_prior}, \cite{ibrahim_2015}, \cite{normalized_power_prior}. The power prior for the target parameters is proportional to an initial prior multiplied by the source data likelihood raised to a fractional power. The fractional power serves to diminish the information provided by the source data likelihood. In our transfer learning setting, the joint prior for $(\theta_C, \theta_T)$ in the target model is given by 
\begin{equation}\label{power_prior_eq}
\pi_{a_0}(\theta_C, \theta_T \mid X_S) \propto p(X_S \mid \theta_C)^{a_0} \pi(\theta_C) \pi(\theta_T),
\end{equation}
where the strength of information transfer ranges between no transfer at $a_0 = 0$ to "full" transfer at $a_0 = 1$. In the latter case, the source data are given equal weight to those in the target domain. This setup 
generalizes the partial borrowing power prior of \cite{partial_borrowing_power_prior}, where the source model parameters are a subset of those used in the target domain. 

Several appealing theoretical properties of the power prior were established in \cite{optimality_power_prior} for the case when all the parameters are shared. In that case the posterior for $\theta$ reduces to 
\begin{equation}\label{power_prior_common}
    \pi_{a_0}(\theta \mid X_T, X_S) \propto p(X_T \mid \theta) p(X_S \mid \theta)^{a_0} \pi(\theta),
\end{equation}
where $\theta$ determines the distribution of both source and target data. Ibrahim \textit{et al.} \cite{optimality_power_prior} show that for a fixed $a_0$, \eqref{power_prior_common} minimizes the weighted sum of Kullback–Leibler (KL) divergences between the posterior with no information transfer and one with full information transfer, i.e. 
\begin{align*}\label{KL_power_prior}
    \pi_{a_0}&(\theta \mid X_T, X_S) =\\ 
&=\argmin_g\{(1-a_0)KL(g \mid \mid f_0) + a_0 KL(g \mid \mid f_1)\},
\end{align*}
where $f_0$ and $f_1$ are probability densities given by 
$$f_0(\theta) = \pi_{0}(\theta \mid X_S, X_T) \propto p(X_T \mid \theta) \pi(\theta)$$
and 
$$f_1(\theta) = \pi_{1}(\theta \mid X_S, X_T) \propto p(X_T \mid \theta) p(X_S \mid \theta)\pi(\theta).$$

Just like in the other transfer learning approaches, choosing the right amount of information to be transferred---in this case governed by the value of $a_0$---is a key challenge. One approach is to treat $a_0$ as fixed and perform sensitivity analysis over a set of values which ideally should include $a_0 = 0$ and $a_0 = 1$, as recommended by \cite{ibrahim_2015}. In  generalized linear models (GLMs) the choice of $a_0$ can be better informed with the help of model selection criteria such as those proposed in \cite{optimality_power_prior}, \cite{ibrahim_2015}, \cite{a0_criterion_1}, and \cite{a0_criterion_2}. Ibrahim \textit{et al.} \cite{optimality_power_prior} propose a 
penalized likelihood-type criterion (PLC) that chooses $a_0 \in (0,1]$ to be the minimizer of 
\begin{equation*}\label{PLC}
     -2\log \int p(X_T \mid \theta)p(X_S \mid \theta)^{a_0} \pi(\theta) d\theta + \frac{\log(n_S)}{a_0},
\end{equation*}
where $n_S$ is the sample size of the source dataset.

Alternatively, we can treat $a_0$ as random and in turn assign it a prior distribution. We can either directly define the joint prior for $(\theta, a_0)$ as in \cite{ibrahim_2015}, i.e.

\begin{equation}\label{joint_power_prior}
    \pi(\theta, a_0 \mid X_S) \propto p(X_S \mid \theta)^{a_0} \pi(\theta)\pi(a_0),
\end{equation}
or the \textit{normalized power prior} as in \cite{normalized_power_prior}
\begin{align}\label{normalized_power_prior}
    \pi(\theta, a_0 \mid X_S) &= \pi(\theta \mid X_S, a_0)\pi(a_0)  \\
    &= \frac{p(X_S \mid \theta)^{a_0} \pi(\theta)}{\int p(X_S \mid \theta')^{a_0} \pi(\theta') d\theta'} \pi(a_0). \nonumber
\end{align}
The normalized power prior first specifies a marginal prior for $a_0$ and then a conditional prior for $\theta$ given $a_0$. 

Taking $\pi(a_0)$ to be a beta or Dirichlet distribution depending on the number of source domains is a natural choice, with theoretical support proved in \cite{optimality_power_prior} under fixed $a_0$ that extend to the random $a_0$ case under \eqref{joint_power_prior}. Other priors with an appropriate support, such as gamma or Gaussian truncated to $[0,1]$, can also be utilized \cite{power_prior_1}. However, it is not clear how the data inform about an appropriate value for $a_0$, since $a_0$ is not a traditional  parameter. It may be that this approach can be used to represent prior uncertainty in $a_0$ but will not adapt to information in the data to concentrate on the optimal amount of borrowing from the source data. 



\subsection{Hierachical models and random effects}

The approaches mentioned in the previous section rely on sharing parameters in the likelihood specification for source and target data. Alternatively, we can allow the parameters of the source and target data models to differ, instead imposing the assumption that they come from a jointly specified or identical prior distribution acting as   a bridge for information flow between the domains.
As a simple example, consider the Gaussian linear model. Let the datasets 
$(\bo{X}_1, \bo{y}_1), \ldots, (\bo{X}_K, \bo{y}_K)$ denote the source data and $(\bo{X}_0, \bo{y}_0)$ be the target data, where $\bo{X}_d \in \mathbb{R}^{n_d\times p}$ and  $\bo{y}_d \in \mathbb{R}^{n_d}$  for $d \in \{0,1,\ldots ,K\}$. Under this model we assume that
\begin{equation}\label{normal_hier}
    \bo{y}_d = \bo{X}_d\bo{\beta}_d + \bo{\epsilon}_d,\quad \bo{\epsilon}_d {\sim} \mathcal{N}(0, \sigma_d^2 I_{n_d}),
\end{equation}
with the prior on the coefficients given by $\bo{\beta}_d \sim \mathcal{N}(\bo{\mu}, \bo{\Sigma})$  for $d \in \{0,1,\ldots ,K\}$. Here, the domain-specific parameters $\bo{\beta}_d$ 
are drawn from a common prior distribution $\mathcal{N}(\bo{\mu}, \bo{\Sigma})$, which is often referred to as a random effects distribution. Model \eqref{normal_hier} is a common type of hierarchical regression model for data nested within groups (domains in our terminology).
Data from all the domains are used to inform the random effects mean $\bo{\mu}$ and covariance $\bo{\Sigma}$, inducing borrowing of information.

We can either treat $\sigma_0, \sigma_1, \ldots, \sigma_K$, $\bo{\mu}$, and $\bo{\Sigma}$ as fixed, taking a frequentist approach to inference, or specify hyperpriors for them to obtain a Bayesian hierarchical model. In either case, the random effects covariance $\bo{\Sigma}$ controls how much information transfer there is, analogously to $a_0$ in the power prior approach. Large covariance implies less shrinkage of the $\bo{\beta}_d$ values towards the random effects mean $\bo{\mu}$. In practice, the prior for the random effects mean and covariance will be updated based on information in the data about the variability in the regression coefficients across domains. 

For fixed $\sigma_0, \sigma_1, \ldots, \sigma_K$, $\bo{\mu}$, and $\bo{\Sigma}$, 
\cite{power_prior_hierarchical_models} showed a direct analytic relationship between $\bo{\Sigma}$ and the tuning parameter $a_0$ in the power prior approach, establishing duality between these methods for the Gaussian linear model. 


\subsection{Shared latent space}\label{shared_latent}

Rather than imposing shared parameters on data generating processes for source and target domains, whether explicitly in the likelihoods or at higher levels in a hierarchical model, we can also specify or seek to learn a shared latent space. This approach can be particularly useful in more complex datasets with large numbers of dimensions. 

\subsubsection{Factor analysis}

In the Bayesian context many such examples can be found in the factor analysis literature. Under the classical factor model specification outlined in \cite{west_factor_models} the $i$-th observation $\bo{y}_i \in \mathbb{R}^{p}$ is given by

\begin{equation*}\label{class_fm_first_spec}
    \bo{y}_i = \bo{\Lambda} \bo{\eta}_i + \bo{\epsilon}_i,
\end{equation*}
where $\bo{\eta}_i \overset{\mathrm{iid}}{\sim} \mathcal{N}(\bo{0}, \bo{I}_q)$ are the vectors of latent factors, $\bo{\Lambda} \in \mathbb{R}^{p \times q}$ is the factor loading matrix, $\bo{\epsilon}_i \iid \mathcal{N}(\bo{0}, \bo{\Delta})$ are random noise terms with $\bo{\Delta} = \textnormal{diag}(\delta_1^2, \dots , \delta_p^2)$, and $\bo{\eta}_i$, $\bo{\epsilon}_j$ are independent for any $i, j$. It is commonly assumed that $q \ll p$, i.e. the high dimensional data can be explained using a latent structure of much lower dimensional factors. This model can be equivalently written as a Gaussian distribution with a constrained covariance structure, i.e. 
\begin{equation*}\label{class_fm}
    \bo{y}_i \iid \mathcal{N}(\bo{0}, \bo{\Sigma}), \quad \bo{\Sigma} = \bo{\Lambda}\bo{\Lambda}^T + \bo{\Delta}.
\end{equation*}
The mean-zero assumption on $\bo{y}_i$ comes from the standard practice of centering the data and does not limit the generality of the model.

In \cite{de_vito_2019} and \cite{de_vito_2021} this setup is generalized to the situation with data coming from multiple domains by letting
\begin{equation*}\label{fm_multiple_domains}
    \bo{y}_{k,i} \iid \mathcal{N}(\bo{0}, \bo{\Sigma}_k), \quad \bo{\Sigma}_k = \bo{\Lambda}\bo{\Lambda}^T + \bo{\Phi}_k\bo{\Phi}_k^T + \bo{\Delta}_k,
\end{equation*}
where $\bo{\Delta}_k = \textnormal{diag}(\delta_{k,1}^2, \dots , \delta_{k,p}^2)$ is the error variance matrix, $\bo{\Lambda} \in \mathbb{R}^{p \times q}$, $\bo{\Phi}_k \in \mathbb{R}^{p \times q_k}$ for domain $k = 1, \dots, K$. 
Here $\bo{\Phi}_k\bo{\Phi}_k^T$ accounts for the domain-specific dependencies between the datapoints and $\bo{\Lambda}\bo{\Lambda}^T$ is the underlying shared covariance structure which allows for information transfer between domains.

Analogously to the single domain case above, this model has the equivalent representation
\begin{align}\label{fm_multiple_domains_equiv}
    \bo{y}_{k,i} = \bo{\Lambda}\bo{\eta}_{k,i} + \bo{\Phi}_k \bo{\zeta}_{k,i} + \bo{\epsilon}_{k,i},
\end{align}
with
\begin{align*}
    \bo{\eta}_{k,i} \iid \mathcal{N}(\bo{0}, \bo{I}_{q}), \hspace{0.25cm} \bo{\zeta}_{k,i} \iid \mathcal{N}(\bo{0}, \bo{I}_{q_k}), \hspace{0.25cm}  \bo{\epsilon}_{k,i} \iid \mathcal{N}(\bo{0}, \bo{\Delta}_k), \nonumber
\end{align*}
where $\bo{\eta}_{k,i}$ is a latent factor in the $q$ dimensional shared subspace, $\bo{\zeta}_{k,i}$ are $q_k$ dimensional domain-specific latent factors and $\bo{\epsilon}_{k,i}$ are error terms. Thus, $\bo{\Lambda}$ is the shared factor loading matrix and $\bo{\Phi}_k$ are $p \times q_k$ domain-specific factor loading matrices. 
In this model the transfer of knowledge between domains occurs through information borrowing in the estimation of $\bo{\Lambda}$.

The above model allows for a lot of flexibility between the domains, but  suffers from an identifiability issue known as information switching: the data can be fitted equally well with the shared columns in factor loading matrices transferred from $\bo{\Lambda}$ to $\bo{\Phi}_k$'s. De Vito \textit{et al.} \cite{de_vito_2019}, \cite{de_vito_2021} solve this problem by restricting the augmented matrix $\begin{matrix}[\bo{\Lambda} & \bo{\Phi}_1 & \dots  & \bo{\Phi}_K ] \end{matrix}$ to be lower-triangular. One limitation of this approach is that it imposes an ordering on the domains.
Often when we have multiple source domains there is no natural ordering between them and hence this approach would not be preferred in such a scenario. 

A recent paper proposes a different solution to the problem of information switching by restricting the factor loading matrices to be linear transforms of the shared factor loading matrix and imposing a shared covariance of error terms between the domains \cite{chandra2023}. That is, they assume $\bo{\Phi}_k = \bo{\Lambda}\bo{A}_k$, where $\bo{A}_k \in \mathbb{R}^{q \times q_k}$ and $\bo{\Delta}_k = \bo{\Delta} = \textnormal{diag}(\delta_{1}^2, \dots , \delta_{p}^2)$ for every $k = 1, \dots, K$ in \eqref{fm_multiple_domains_equiv}. The authors show that under any non-degenerate continuous prior on $\bo{A}_k$ the information switching does not occur almost surely provided that $\sum_{k=1}^K q_k \leq q$. 

This result provides some guidance for choosing the dimensions of shared and domain-specific latent spaces since these are not known in most practical applications. These dimensions influence the amount of information being transferred between domains since larger values of $q_1, \dots, q_K$ give more flexibility to the domain-specific latent features, thus reducing the influence of shared latent factors. One approach to choosing these dimensions would be to put priors on $q_1, \dots, q_K$ and $q$ and use reversible-jump algorithms outlined in \cite{green_1995}. However, such algorithms can be computationally prohibitive.

Chandra \textit{et al.} \cite{chandra2023}  provide an alternative solution by fixing $q, q_1, \dots, q_K$ at an upper bound, and then utilizing appropriate priors to shrink the excess columns in $\bo{\Lambda}, \bo{A}_1, \dots, \bo{A}_K$. Specifically, they obtain approximate singular values and eigenvectors of the pooled dataset via the augmented implicitly restarted Lanczos bidiagonalization \cite{Lanczos} and choose the smallest $\hat{q}$ which explains 95\% of the variability in the data. Then they fix $\hat{q_k} = \hat{q} / K$ for $k = 1, \ldots, K$ to ensure that information switching will not occur.
The strength of information transfer between domains is thus directly influenced by the amount of shrinkage induced by the priors on the factor loading matrices. In  \cite{chandra2023}, the fixed priors $\textnormal{vec}(\bo{\Lambda}) \sim \textnormal{DL}(1/2)$ and $a_{k,i,j} \iid \mathcal{N}(0,1)$ are used, where $a_{k,i,j}$ is the $(i,j)$-th entry of $\bo{A}_k$ and DL denotes the Dirichlet-Laplace distribution \cite{DL_prior}. Thus a possible extension in this line of work would be to propose some methods for choosing these hyperparameters in an adaptive way depending on how related the domains are. 


\subsubsection{Mixture models}

Latent space models are commonly used as a dimensionality reduction tool, including when dealing with non-standard data structures such as networks \cite{hoff_2008}, \cite{bu2023inferring}. As a vignette showing how mixture models can be used in combination with latent space models for flexible transfer learning, we focus on the approach proposed in \cite{networks_latent}. They were motivated by data on brain networks for individuals in different groups.

Specifically, given $n$ observed networks each belonging to one of $K$ groups and consisting of $V$ labelled vertices, denote the $i$-th network together with its group label as $\{y_i, \bo{\mathcal{L}}(\bo{A}_i)\}$, where $\bo{A}_i \in \{0,1\}^{V\times V}$ is the adjacency matrix and $\bo{\mathcal{L}}(\bo{A}_i) \in \{0,1\}^{V(V-1)/2}$ denotes the lower triangular entries 
$$(A_{i[2,1]}, \dots , A_{i[V,1]}, A_{i[3,2]}, \dots , A_{i[V,2]}, \dots , A_{i[V,V-1]})^T $$ 
of $\bo{A}_i$. We discard the main diagonal and the upper triangular part of $\bo{A}_i$ since the network is an undirected graph and self-relationships of the nodes are not of interest. In \cite{networks_latent} subjects fall into a low and high creativity group, so we have $K=2$ domains. The network representation $\bo{\mathcal{L}}(\bo{A})$ conditional on the group membership $y$ is modeled as
\begin{align*}\label{network_model}
    \mathbb{P}(\bo{\mathcal{L}}&(\bo{A}_i) = \bo{a} \mid y = k) =\\ 
    &=\sum_{h=1}^H \nu_{h}^{(k)} \prod_{l=1}^{V(V-1)/2} \left(\pi_l^{(h)}\right)^{a_l}\left(1 - \pi_l^{(h)}\right)^{1-a_l}
\end{align*}
for any $\bo{a} \in \{0,1\}^{V(V-1)/2}$ with the probability vector 
$$\bo{\pi}^{(h)} = \left(\pi_1^{(h)}, \dots , \pi_{V(V-1)/2}^{(h)}\right)^T \in (0,1)^{V(V-1)/2}$$
in the $h$-th mixture component given by
\begin{equation*}\label{graph_factorization}
    {\pi}^{(h)}_l = \left[1 + \exp(-{Z}_l - {D}^{(h)}_l)\right]^{-1},
\end{equation*}
with
\begin{equation*}
\bo{D}^{(h)} = \bo{\mathcal{L}}(\bo{X}^{(h)}\bo{\Lambda}^{(h)}\bo{X}^{(h)T}), \quad h = 1, \dots, H,
\end{equation*}
where $\bo{X}^{(h)} \in \mathbb{R}^{V\times R}$, $\bo{\Lambda}^{(h)} = \textnormal{diag}(\lambda_1^{(h)}, \dots , \lambda_R^{(h)})$ with $\lambda_1^{(h)}, \dots , \lambda_R^{(h)} \geq 0$, and $\bo{Z} \in \mathbb{R}^{V(V-1)/2}$. 

The model supposes that there are $H$ different brain structure ``types''. The probability of an edge between the $l$-th pair of brain regions follows a logistic model having an intercept $Z_l$ characterizing the baseline log odds of a connection and a low-rank deviation that differs according to the individual's brain type. To enable information transfer across the creativity groups (domains), the model assumes the brain structure types do not differ across the groups (referred to as ``common atoms'' in the mixture modeling literature). However, the proportion of individuals 
having brain type $h$, $\nu_{h}^{(y)}$, does differ across domains $y=1,\ldots,K$. 

Although the goal in \cite{networks_latent} was inference on group differences, this model can be used directly for transfer learning from  source domains to a target domain, with the source data enabling more accurate estimation of the shared network types.
In addition, the baseline log-odds of an edge between each pair of nodes is also shared across the groups, leading to information sharing about common topological properties of the graphs, 
including block structures, homophily behaviors and transitive edge patterns \cite{hoff_2008}. 
An important application would be to transferring information from large brain imaging repositories, such as the Human Connectome Project (HCP) and UK Biobank, to small neuroimaging studies in targeted populations.

In \cite{lock_dunson_shared_kernel} shared kernels are used to model complex distributions of multiple variables across different domains. The motivating applications are studies investigating how DNA methylation profiles vary according to cancer subtype. For samples $i=1,\ldots,n$, data consist of $\bo{x}_i = (x_{i1},\ldots,x_{ip})^T$ with $x_{ij}$ denoting the methylation level at site $j$, for $j=1,\ldots,p$ with $p$ very large (e.g., $p=450,000$) and $y_i \in \{1,\ldots,K\}$ denoting the group membership. The density of the data in group $k$ for the $j$th variable is 
$f_j^{(k)}(\cdot)=\sum_{h=1}^H \nu_{jh}^{(k)}\mathcal{K}(\cdot;\bo{\theta}_h)$, with $\mathcal{K}(\bo{\theta})$ a family of densities parameterized by $\bo{\theta}$ and $\nu_{jh}^{(k)}$ a probability weight on kernel $h$ specific to site $j$ and group $k$.

Although the motivation in \cite{lock_dunson_shared_kernel} is 
testing for differences in methylation between different groups, the proposed approach can be directly applied to transfer learning focused on inferring the marginal densities of very high-dimensional data within a particular domain.
The data from all the domains are used in inferring the shared kernel parameters $\bo{\theta}_1,\ldots,\bo{\theta}_H$ and further borrowing of information occurs through a hierarchical model for the weights $\{ \nu_{jh}^{(k)} \}$. Even in using shared kernels, this approach  allows highly flexible differences in distribution across the groups.

There is a rich literature on alternative Bayesian mixture models for borrowing of information across grouped data, while also allowing distinct characteristics of each group.
Suppose we let $\bo{x}_i$ denote feature data for subject $i$ with $y_i \in \{1,\ldots,K\}$ denoting the subject's group membership. Then, a common approach is to incorporate subject-specific parameters $\bo{\theta}_i$ within the likelihood function for $\bo{x}_i$ and then let $\bo{\theta}_i \sim P_{y_i}$, 
with the collection of group-specific random effects distributions $(P_1,\ldots,P_K) \sim \Pi$ given an appropriate prior. Popular choices of $\Pi$ include the hierarchical Dirichlet process (HDP) \cite{Teh_2006} and nested Dirichlet process (NDP) \cite{rodriguez_2008}, both of which fall within the broad class of hierarchical processes \cite{hier_proc}. These approaches 
characterize each $P_k$ as almost surely discrete while incorporating statistical dependence between $P_k$ and $P_l$ for all $k \neq l$, leading to dependence in clustering.


Alternatively, analogously to the multi-group factor models of \cite{de_vito_2019}, \cite{de_vito_2021}, and \cite{chandra2023}, 
 M\"uller \textit{et al.} \cite{Muller_2004} modeled the group-specific random effects distributions $P_k$ as a mixture of a common cross-group distribution $P_0$ and group-specific distributions $Q_k$, with a hyperprior chosen for the mixture weight on $P_0$ to allow data adaptivity. This approach, and the above approaches, assume {\em a priori} exchangeability across the groups. In the future, it will be interesting to adapt these approaches and develop appropriate extensions explicitly targeting the transfer learning case in which one domain is the particular focus. A relevant recent advance is the graphical Dirichlet process of Chakrabarti \textit{et al.} \cite{chakrabarti2023graphical}, which incorporates a known directed acyclic graph (DAG) characterizing the dependence structure across the groups.



\subsection{Network transfer}\label{network_transfer}

As an alternative to viewing each source domain as providing exchangeable information about the target domain {\em a priori}, there is often expert knowledge about directed relationships between the different domains. Incorporating a network of relationships among the domains in transfer learning is referred to as {\em network transfer}, as opposed to {\em direct transfer}.

\begin{figure}[]
    \centering
\begin{tikzpicture}[node distance={17mm}, thick, main/.style = {draw, circle}] 
\node[main] (1) {$\mathcal{L}_{T}$}; 
\node[main] (2) [above right of=1] {$\mathcal{L}_{S_1}$};
\node[main] (3) [below right of=1] {$\mathcal{L}_{S_2}$}; 
\node[main] (4) [below left of=1] {$\mathcal{L}_{S_3}$};
\node[main] (5) [above left of=1] {$\mathcal{L}_{S_4}$};
\draw[->] (2) -- (1);
\draw[->] (3) -- (1);
\draw[->] (4) -- (1);
\draw[->] (5) -- (1);
\end{tikzpicture} 
\hspace{0.4cm}
\begin{tikzpicture}[node distance={17mm}, thick, main/.style = {draw, circle}]  
\node[main] (1) {$\mathcal{L}_{S_1}$}; 
\node[main] (2) [above right of=1] {$\mathcal{L}_{T}$};
\node[main] (3) [below right of=1] {$\mathcal{L}_{S_3}$}; 
\node[main] (4) [above right of=3] {$\mathcal{L}_{S_2}$};
\node[main] (5) [below right of=4] {$\mathcal{L}_{S_4}$};
\draw[->] (1) -- (2);
\draw[->] (4) -- (2);
\draw[->] (1) -- (2);
\draw[->] (3) -- (4);
\draw[->] (5) -- (4);
\draw[->] (3) -- (1);
\end{tikzpicture} 
    \caption{Direct transfer (left) and network transfer learning (right). In direct transfer all the source learners are used directly in supporting the training of the target learner, whereas in network transfer we can have a more complex structure with some of the source learners supporting other source learners rather than the target learner directly.} 
    \label{fig:enter-label}
\end{figure}
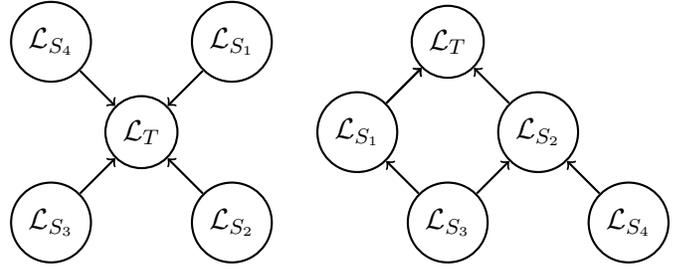

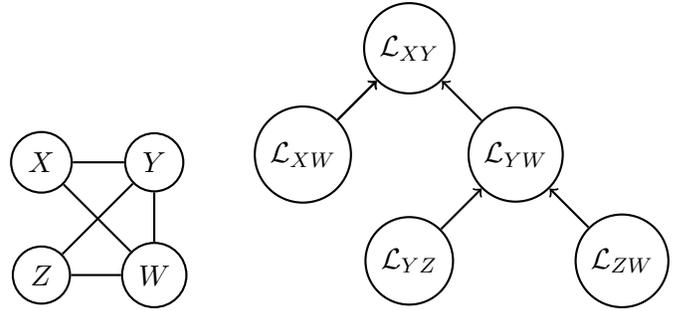
\begin{figure}[]
    \centering
\begin{tikzpicture}[node distance={15mm}, thick, main/.style = {draw, circle}] 
\node[main] (1) {$X$}; 
\node[main] (2) [right of=1] {$Y$};
\node[main] (3) [below of=1] {$Z$}; 
\node[main] (4) [below of=2] {$W$};
\draw (1) -- (2);
\draw (2) -- (4);
\draw (3) -- (4);
\draw (1) -- (4);
\draw (2) -- (3);
\end{tikzpicture} 
\hspace{0.7cm}
\begin{tikzpicture}[node distance={20mm}, thick, main/.style = {draw, circle}]  
\node[main] (1) {$\mathcal{L}_{XW}$}; 
\node[main] (2) [above right of=1] {$\mathcal{L}_{XY}$};
\node[main] (3) [below right of=1] {$\mathcal{L}_{YZ}$}; 
\node[main] (4) [above right of=3] {$\mathcal{L}_{YW}$};
\node[main] (5) [below right of=4] {$\mathcal{L}_{ZW}$};
\draw[->] (1) -- (2);
\draw[->] (4) -- (2);
\draw[->] (1) -- (2);
\draw[->] (3) -- (4);
\draw[->] (5) -- (4);
\end{tikzpicture} 
    \caption{Evidence network for the treatment comparison (left) and the network transfer of information between the associated learners in the meta analysis (right).} 
    \label{fig:network_transfer_fig}
\end{figure}

In Bayesian network meta-analysis \cite{network_meta_2004} the goal is often to compare the efficacy of a pair of treatments based on multiple studies, some of which may involve arms with other treatments. Let $W,X,Y,Z$ be four available treatments among which we want to compare the efficacy of $X$ and $Y.$ Suppose that we have the dataset $\mathcal{D}_{XY}$ formed based on studies comparing $X$ and $Y$ and that we also have access to datasets $\mathcal{D}_{XW}$, $\mathcal{D}_{YW}$, $\mathcal{D}_{YZ}$, and $\mathcal{D}_{WZ}$ comparing, respectively $X$ to $W$, $Y$ to $W$, $Y$ to $Z$, and $W$ to $Z$. We may have several different trials for certain of these comparisons. The knowledge extracted from the trials comparing other treatments can be used to indirectly improve the analysis of the $X$ vs $Y$ trials. Figure \ref{fig:network_transfer_fig} shows a graph representing the observed comparisons between the treatments, sometimes referred to as the evidence network \cite{bnma_2006}, and the associated network transfer graph. 

The general framework for Bayesian network meta-analysis is outlined in \cite{bnma_2006}. Denote the observed mean difference in the efficacy of treatments $k$ and $l$ in study $i$ by $\delta_{i,k,l}$ and the baseline difference in efficacy between treatments $k$ and $l$ by $d_{k,l}$. We refer to $d_{k,l}$ as effect parameters. In \cite{bnma_2006} they are divided into basic parameters $\bo{d}_b$ and functional parameters $\bo{d}_f$. Any set of effect parameters can be treated as basic parameters if the edges associated with them create a spanning tree of the evidence network. The functional parameters are the remaining effect parameters. 

Network meta analysis assumes functional parameters can be represented as linear functions of basic parameters, i.e. $\bo{d}_f = \bo{F}\bo{d}_b$ for some matrix $\bo{F}$, which is referred to as \textit{evidence consistency}. Usually, these relations take the form $d_{j,k} = d_{j,l} - d_{k,l}$ for any treatments $j,k,l$. In our example, we can choose ${d}_{X,W},  {d}_{Y,W}, {d}_{Z,W}$ to be the basic parameters and then relate the functional parameters to them via $d_{X,Y} = d_{X,W} - d_{Y,W}$ and $d_{Y,Z} = d_{Y,W} - d_{Z,W}$. Leveraging this assumption is analogous to utilizing the shared parameter strategy outlined in section \ref{shared_params_section}. We can use these identities to increase the precision of estimation of $d_{Y,W}$ which in turn, together with the estimates of $d_{X,W}$, can increase the precision of the estimation of $d_{X,Y}$. This is represented by the network transfer in Figure \ref{fig:network_transfer_fig}.

The linear relationship $\bo{d}_f = \bo{F}\bo{d}_b$ can be used to model the vector of observed differences in treatments $\bo{\delta}$ conditionally on $\bo{d}_b$ and the covariance of $\bo{\delta}_b$, denoted by $\text{Cov}(\bo{\delta}_b) = \bo{V}_b$. Using the Gaussian distribution $\bo{\delta} \sim \mathcal{N}\left(\left(\bo{d}_b^T, \bo{d}_b^T\bo{F}^T\right)^T, \bo{V}\right)$ is standard \cite{network_meta_2004}, \cite{network_meta_Dias2011}, \cite{leblanc2023}, where 
$$
\bo{V} = 
\begin{pmatrix}
\bo{V}_b & \bo{V}_b\bo{F}^T \\
\bo{F}\bo{V}_b^T & \bo{FV}_b\bo{F}^T
\end{pmatrix}.
$$
This prior can be incorporated within a hierarchical model for the individual observations in each study. Further borrowing of information can be facilitated through placing a common random effects distribution on the basic treatment effect parameters as in \cite{leblanc2023}.

Additional flexibility in information transfer can come from allowing violations in evidence consistency. Lu \textit{et al.} \cite{bnma_2006} provide such a framework via 
$\bo{d}_f = \bo{F}\bo{d}_b + \bo{w}$, where $\bo{w}$ represents inconsistencies between  studies. In our example 
$$d_{X,Y} = d_{X,W} - d_{Y,W} + w_{X,Y,W}$$ 
and 
$$d_{Y,Z} = d_{Y,W} - d_{Z,W} + w_{Y,Z,W}.$$ 
The inferred size of $\bo{w}$ directly measures how related the domains are and determines how much information transfer should occur between them. 
There can be various sources of inconsistencies between the pairwise comparisons. They can stem from limitations in the design of individual studies and from changes in the baseline efficacy of treatments over time, for example due to increasing antibiotic resistance. This problem has recently been addressed in \cite{leblanc2023} where the basic parameters are assumed to vary over time according to a Gaussian process. Thus, the information transfer between domains is corrected for the times at which the associated datasets were collected.



Often the appropriate transfer network joining the domains is not known and needs to be inferred. One can take a brute-force approach to select the best transfer network under some quality measure by checking every possible graph.  However, this approach quickly becomes intractable as the number of domains grows with millions of possible transfer networks on just eight domains. Zhou \textit{et al.} \cite{multiple_learners_graph} provide a greedy algorithm which starts from the target learner and at each step includes a source learner yielding the highest conditional marginal likelihood for the target task. Specifically, suppose we have learners $\mathcal{L}_1, \dots ,\mathcal{L}_K$ operating on datasets $\bo{D}^{(1)} = (\bo{y}^{(1)}, \bo{X}^{(1)}), \dots , \bo{D}^{(K)} = (\bo{y}^{(K)}, \bo{X}^{(K)})$, respectively, where $\mathcal{L}_1$ is the target learner. Let ${G} = (V,E)$ be the (connected) network transfer graph with $V = \{1, \dots, K \}$. Let $\bo{\theta}_1, \dots , \bo{\theta}_K$ be {the} parameters in $\mathcal{L}_1, \dots , \mathcal{L}_K$, where for every $(i,j) \in E$ there exist subvectors $\bo{\theta}_{i,\mathcal{C}_i}$ and $\bo{\theta}_{j,\mathcal{C}_j}$ of, respectively, $\bo{\theta}_i$ and $\bo{\theta}_j$ which are restricted to be equal (shared parameter approach). Then at each step, given the chosen set of learners $Q \subset V$, which is referred to as the \textit{linkage set}, let $N_G(Q)$ be the set of neighbors of $Q$, consisting of all learners adjacent to at least one learner in $Q$. We then select a new learner $j^*$ to be added to $Q$ via 
\begin{equation}\label{learner_select_rule}
    j^* = \argmax_{j \in N_G(Q)} p(\cup_{k \in Q} \bo{y}^{(k)} \mid \cup_{k \in Q} \bo{X}^{(k)}, \bo{D}^{(j)}).
\end{equation}
The algorithm terminates once adding a new learner no longer increases the conditional likelihood in \eqref{learner_select_rule}. The complexity of this algorithm is $O(K^2)$ under the assumption that the conditional likelihood in \eqref{learner_select_rule} can be obtained in constant time. This can be further reduced to $O(K \log K)$ if the likelihood computation is parallelized between the learners in $Q$.   
Zhou \textit{et al.} \cite{multiple_learners_graph} provide theoretical guarantees for the recovery of the optimal transfer subnetwork of $G$. 




Having explored a variety of Bayesian approaches to transferring information between domains in a flexible manner, we now discuss which among these are applicable to particular types of transfer learning problems depending on (i) feature spaces of source and target domains; (ii) the availability of labels and samples in both source and target datasets. We note that there is an immense Bayesian literature providing relevant models for transfer learning that we do not mention above. Instead, we have chosen to highlight some approaches that we find particularly interesting and useful.





    
    \section{Transfer with Common vs Overlapping Variables}\label{interp_intra_sec}
    
    Following criterion (i), transfer learning problems can be dichotomized based on whether or not the observations in the source and target domains live in the same feature spaces. We will refer to the former case as \textit{common variables} and to the latter as \textit{overlapping variables transfer}.
    This classification often determines which of the approaches presented in Section \ref{methods_sec} are appropriate or even feasible to use.

    \subsection{Common variables transfer}
    Common variables transfer, also known as as {homogeneous transfer} \cite{opt_BTL}, occurs when the source and target data and labels have the same meaning in the different domains, but may follow different distributions. For example, the same variables are collected for each of the study subjects but subjects in different groups may have considerably different attributes; hence, the distribution of the variables being collected may vary across groups. In this common variables transfer case, it typically makes sense to define the same form of likelihood for the data in each domain, though there may be systematic differences in the parameters. All of the methods described in Section \ref{methods_sec} can be applied directly to common variables problems.

Examples of common variables transfer include 
clinical studies with patients divided according to their health status or subtype of disease. Researchers are often interested in improving the accuracy of inference and predictions for a particular group of patients by utilizing the information gathered from the other groups or from healthy individuals. This can be especially useful when dealing with rare diseases where it is often difficult to collect measurements on a large sample of affected patients. In \cite{cancer_rna} and \cite{osti_10096293}, the authors use RNA sequencing datasets for different types of lung, kidney, head and neck cancer as source domains in order to improve the accuracy of subtype identification for particular types of lung cancer. Bayesian models borrowing strength across classes and types of cancer have been applied in other contexts, including survival analysis \cite{pan_cancer_surv}, \cite{pan_cancer_surv2}, and protein network inference \cite{prot_network}, \cite{prot_network2}.

\subsection{Overlapping variables transfer}

This case is more challenging as the source and target datasets do not consist of measurements on exactly the same variables for different study subjects. In order for transfer learning to apply, there has to be something in common across the domains. A typical setting is when the domains have overlapping but not completely identical sets of variables. For example, there may be a common focus across the domains in studying the impact of key predictors of interest on a response, measured under different covariates.

For regression or classification models, the coefficients for the key predictors are not directly comparable across models adjusting for different covariates. Hence, shared parameter models are not appropriate. Nonetheless, it may 
make sense to assume that the domain-specific coefficients for the key predictors are drawn from a common random effects distribution, thereby enabling borrowing of information. Shared latent space models are even more natural in this case.
By jointly modeling the response, key predictors, and covariates as conditionally independent given latent factors, we induce a parsimonious {\em latent factor regression/classification} model \cite{Carvalho2008}, \cite{ferrari2021}. Multi-study variants of such models can seamlessly handle cases in which the covariates differ across domains. The multi-group variants of Bayesian mixture models discussed in Section \ref{methods_sec} similarly apply for the joint distributions of key predictors, covariates and response(s) to induce flexible transfer learning.


  
Bayesian methods enjoy a key advantage in this context, in their ability to rely on joint latent feature models to transfer information across domains with partially overlapping variables. 
The majority of non-Bayesian approaches  such as deep transfer learning (\cite{deep_tl_1}, \cite{deep_tl_survey}, \cite{deep_tl_imaging}, \cite{deep_tl_tzeng}, \cite{deep_tl_luo}, \cite{deep_tl_long}) rely on both domains having observations in the same feature space for pre-training and fine-tuning the learners. We detail some examples of  Bayesian transfer with overlapping variables below.

\subsubsection{Multi-study latent factor regression}

Recall the multi-study latent factor model in equation \eqref{fm_multiple_domains_equiv}. Previously, we considered the observed data on subject $i$ in study (domain) $k$, $\bo{y}_{k,i}$, to be $p$-dimensional, with $p$ fixed across subjects and domains. To generalize this, we instead consider the $p$-dimensional data $\bo{y}_{k,i}$ to be the {\em complete data} for subject $i$ in study $k$ that could have potentially been measured. Then, we define $\bo{m}_{k,i}=(m_{k,i,j},j=1,\ldots,p)^T$ to be the {\em missingness pattern} for the $(k,i)$ subject, with 
$m_{k,i,j}=1$ if the $j$th variable is not observed for that subject and $m_{k,i,j}=0$ otherwise. A variable is missing for a subject if the study they participated in does not collect that variable, or if the study planned to collect that variable but it was not available.

Let $\bo{y}_{k,i}^{(obs)}=\{ y_{k,i,j}, j:m_{k,i,j}=0, j=1,\ldots,p\}$ denote the 
$p_{k,i}=\sum_{j=1}^p (1-m_{k,i,j})$ dimensional observed data vector for subject $(k,i)$.
The Gaussian multi-study latent factor model characterizes the complete data vector as 
$\bo{y}_{k,i} \sim \mathcal{N}(\bo{0}, \bo{\Sigma}_k)$. This in turn induces a 
$p_{k,i}$-dimensional multivariate Gaussian distribution for the observed data vector $\bo{y}_{k,i}^{(obs)}$ having covariance corresponding to the appropriate sub-matrix of $\bo{\Sigma}_k$. In fitting Bayesian multi-study factor models, it is not necessary to impute the missing data. Instead, one can simply 
take into account the differing observed data contributions for each subject in implementing a Gibbs sampler or alternative Markov chain Monte Carlo (MCMC) algorithm for posterior sampling.

This approach can be used for transfer learning about the covariance structure in multivariate data specific to the target domain. Alternatively, when the focus is on regression, one can concatenate outcomes, predictors of interest and covariates together in the $\bo{y}_{k,i}$ data vector for subject $(k,i)$. A Gaussian linear regression model for the outcome given the predictors of interest and covariates can be then obtained directly from the covariance 
$\bo{\Sigma}_k$ using standard multivariate Gaussian theory. This type of approach is straightforward to extend to mixed categorical and continuous data by following the popular approach of linking categorical variables to underlying Gaussian variables.

\subsubsection{Nonlinear and nonparametric extensions}

A limitation of the above multi-study factor analysis model is the assumption of multivariate Gaussianity. It is hence useful to consider extensions that incorporate shared and study-specific latent factors while relaxing these restrictive distribution assumptions, which also imply linear relationships among the variables.

In the single modality case, there is a rich literature on nonlinear factor models. For example, we could let 
\begin{eqnarray}
 \bo{y}_i = f(\bo{\eta}_i) + \bo{\epsilon}_i,\quad \bo{\epsilon}_i \overset{\mathrm{iid}}{\sim} \mathcal{N}(\bo{0}, \sigma^2 \bo{I}_p), \label{eq:npfactor}
\end{eqnarray}
where $\bo{\eta}_i \overset{\mathrm{iid}}{\sim} \mathcal{N}(\bo{0}, \bo{I}_q)$ are the vectors of latent factors and $f(\cdot)$ is an unknown and potentially non-linear function.  Gaussian process latent variable models (GP-LVMs)  place a GP prior on the function $f$ mapping from the latent to ambient space \cite{hier_GP_LVM}, \cite{multiview_GP_LVM}, \cite{song2019harmonized}, \cite{xu2023identifiable}. Alternatively, the popular class of variational autoencoders (VAEs) characterize $f$ using deep neural networks and take a variational approach to inference \cite{moran2022identifiable}, \cite{domain_inv_vae}, \cite{auto_vae}.

While these highly flexible nonlinear latent variable models have exhibited appealing practical performance as black-box models for generating new data that resemble the training data, they are prone to a number of vexing issues in reproducing statistical inferences. One major challenge is the curse of dimensionality resulting from the fact that the function $f$ is an unknown mapping from $q$ to $p$ dimensional space; the space of such functions is immense, necessitating an enormous amount of training data for adequate performance. Furthermore, these models are not identifiable without substantial additional constraints.
Another common problem is referred to as {\em posterior collapse} \cite{post_collapse}, \cite{posterior_collapse}, \cite{post_collapse_2}, in which there is a lack of learning about the latent variables based on the data.
While there have been some attempts at addressing these problems, there remains a lack of practically useful methodology to perform reproducible dimensionality reduction.


The above challenges are exacerbated in considering extensions to the multi-study (transfer learning) case. Hence, we recommend starting with more parsimonious nonlinear latent factor models in future work developing such extensions. One promising point of departure is the recently proposed NIFTY framework of Xu \textit{et al.} \cite{xu2023identifiable}, which lets 
\begin{eqnarray*}
\bo{y}_i & = & \bo{\Lambda}\bo{\eta}_i + \bo{\epsilon}_i,\quad \bo{\epsilon}_i \overset{\mathrm{iid}}{\sim} \mathcal{N}(\bo{0}, \bo{\Sigma}), \nonumber \\
\eta_{ih} & = & g_h(u_{ik_h}),\quad h=1,\ldots,q,
\end{eqnarray*}
where $\bo{\Lambda}$ is a factor loading matrix, 
$\bo{\Sigma}=\mbox{diag}(\sigma_1^2,\ldots,\sigma_p^2)$, and $u_{ik} \stackrel{iid}{\sim}U(0,1)$ for $k=1,\ldots,K \le q$. Each latent factor $\eta_{ih}$ is a transformation of a latent $u_{ik_h}$ via an unknown non-decreasing function $g_h$. The subscript $k_h$ allows the same latent uniforms to be used for multiple factors, inducing dependence. 


This model induces a flexible multivariate density for $\bo{y}_i$ while massively reducing dimensionality relative to model \eqref{eq:npfactor}. In their paper, they provided theory on identifiability, leveraging on pre-training with state-of-the-art nonlinear dimensionality reduction algorithms. They also 
showed excellent performance for a wide variety of complex examples. They were even able to train a realistic generative model for bird songs based on few training examples; audio recordings of bird songs provide an example of massive dimensional data with low intrinsic dimension. NIFTY can exploit the complex low dimensional structure in the data for highly efficient performance. 

In conducting inference for latent variable models, the NIFTY authors noticed a common problem of {\em distributional shift}. In particular, many of the current models assume that the latent variables are iid $\mathcal{N}(0,1)$ or $U(0,1)$. Inferences on the parameters, such as the induced covariance in the Gaussian linear factor model case, critically depend on this assumption holding not just {\em a priori} but also {\em a posteriori}. Xu \textit{et al.} \cite{xu2023identifiable} propose a general approach for solving latent variable distributional shift through forcing the posterior distribution of the latent variables to be very close to iid $U(0,1).$

\subsubsection{Mixture models}

An alternative direction towards building more flexible models for transfer learning, including in the partially overlapping variables case, is to rely on mixture models, building on the developments in Section \ref{shared_latent}. Such models have the advantage of also clustering subjects within the different domains.
In the partially overlapping variables transfer learning case, it is appealing to define a joint model, as motivated above. However, Chandra \textit{et al.} \cite{Clustering_Chandra} recently noted a pitfall of mixture models in high dimensional cases in which the posterior tends to concentrate on trivial clusterings of the observations that place all subjects into one cluster or in singleton clusters.


As a solution in the single domain case, they proposed a latent mixture model formulation that lets \begin{eqnarray}
\bo{y}_i & = & \bo{\Lambda}\bo{\eta}_i + \bo{\epsilon}_i,\quad \bo{\epsilon}_i \overset{\mathrm{iid}}{\sim} \mathcal{N}(\bo{0}, \bo{\Sigma}), \nonumber \\ 
\bo{\eta}_i & \sim & \sum_{h=1}^H \nu_h \mathcal{N}(\bo{\mu}_h, \bo{\Delta}_h), \label{eq:lamb}
\end{eqnarray}
so that a mixture of Gaussians model is used for the latent variables in a linear factor model. They prove that this model solves the above mentioned pitfall in Bayesian clustering in high dimensions. The trick is to model the variation across clusters in a lower dimensional latent space to address the curse of dimensionality.

With this single domain specification as the starting point, there are multiple promising directions forward in terms of extensions to the multiple domain transfer learning case. One possibility is to define a multi-study factor model as in Chandra \textit{et al.} \cite{chandra2023} but instead of assuming Gaussian shared and study-specific latent factors, use Gaussian mixture models to induce a flexible distribution on the latent factors while also producing separate clusters of subjects in each domain with respect to the shared and study-specific components. An alternative is to rely on the model in equation \eqref{eq:lamb} but with domain-specific distributions for the latent factors defined as 
\begin{eqnarray*}
f_k(\bo{\eta}_i) = \int \mathcal{N}(\bo{\eta}_i;\bo{\theta}_i)dP_k(\bo{\theta}_i), 
\end{eqnarray*}
where $\bo{\theta}_i = \{\bo{\mu}_i,\bo{\Delta}_i\}$ are the Gaussian parameters and 
$P_k$ is a mixing distribution on these parameters that is specific to domain $k$.

In the special case in which $P_k=P=\sum_{h=1}^H \nu_h \delta_{\bo{\theta}_h^*}$ with $\delta_{\bo{\theta}}$ a degenerate distribution concentrated at $\bo{\theta}$ we obtain the original model in \eqref{eq:lamb}. However by using the different priors $(P_1,\ldots,P_K)\sim \Pi$ considered in Section \ref{shared_latent} we can allow differences across the domains while borrowing information; further borrowing is achieved through the implicit assumption of a latent space that is shared across domains - this is induced through the use of a common factor loading matrix 
$\bo{\Lambda}$.

\subsubsection{Multiresolution transfer learning}

Closely related to the overlapping variables case is the setting in which data are collected for each domain on related random functions or stochastic processes. For example, let $f_k: \mathcal{T} \to \mathbb{R}$ denote a latent smooth continuously differentiable function for domain $k$, and suppose that we have 
\begin{eqnarray}
y_{k,i} = f_k(t_{k,i}) + \epsilon_{k,i},\quad 
\epsilon_{k,i} \iid \mathcal{N}(0,\sigma^2),\label{eq:FDA}
\end{eqnarray}
with $\bo{y}_k=\{ y_{k,i}, i = 1,\ldots,n_k \}$ the observed data and 
$\bo{t}_k = \{ t_{k,i}, i=1,\ldots,n_k \}$ the observation locations for domain $k$. Often, the observation locations do not line up across domains and certain domains may have lower resolution data than others, with the {\em resolution} referring to the density of the observation points $\bo{t}_k$ over the support 
$\mathcal{T}.$

Model \eqref{eq:FDA} represents a type of {\em functional data analysis} (FDA). In many FDA settings, we observe noisy realizations of a subject-specific function at a finite set of points, but here we are considering the case in which we have one function for each domain and are particularly interested in inference on the function for the source domain. There are many applications in which this type of problem arises. We may have domain-specific regression functions $f_k$ and want to borrow information across domains in a nonparametric regression context without assuming any common parameters. Alternatively, $\bo{y}_k$ may correspond to a domain $k$-specific time series and we want to borrow information across related time series to enhance prediction for a target series. 

A natural Bayesian approach to inference under model \eqref{eq:FDA} is to consider a functional data extension of the hierarchical and random effects modeling approaches highlighted in Section \ref{methods_sec}. For example, one could use a hierarchical GP that lets $f_k \sim \mbox{GP}(f_0, c)$ with $f_0$ in turn given a GP prior \cite{hier_GP_2005}, \cite{hier_GP_2019}. For articles on using GPs in closely related transfer learning settings to that of \eqref{eq:FDA} refer to  \cite{multitask_gp}, \cite{aggregated_data_gp}, \cite{multiresolution_multitask_gp}. These approaches can automatically accommodate the case in which observations are denser in some domains than others.

Wilson \textit{et al.} \cite{multitask_gp} introduce GP regression networks (GPRN) which use latent GPs to transfer information between different continuous time processes. Specifically, given $K$ time series $\bo{y}_1, \dots ,\bo{y}_K$, \cite{multitask_gp} let 
$$\bo{y}_k = \sum_{q=1}^Q \bo{W}_{k,q} \odot \bo{f}_q + \bo{\epsilon}_k,$$ 
where $\odot$ is the Hadamard product, $\bo{f}_q \sim \mathcal{GP}(0, \bo{K}_q^f)$ are latent basis GPs evaluated at the observation times, $\bo{W}_{k,q} \sim \mathcal{GP}(0, \bo{K}_{k,q}^w)$ are domain-specific weights of the latent GPs, and $\bo{\epsilon}_k \sim \mathcal{N}(\bo{0}, \sigma_k^2\bo{I})$ represents measurement noise. The weights $\bo{W}_{k,q}$ determine the strength and structure of information transfer between the domains, analogously to the network of learners introduced in \cite{multiple_learners_graph}. However, unlike in most works in Section \ref{network_transfer}, GPRN allows for the strength of the transfer within the network of learners to vary over time. 

Although \cite{multitask_gp} simplify inference by assuming identical measurement times across domains, \cite{multiresolution_multitask_gp} extend the approach to allow different measurement times and increase flexibility by using deep GPs. By using a shared latent space approach with GPs serving as the basis for the latent space, these models are able to achieve transfer both across resolutions and different classes of observations, corresponding to different air pollutants in the application presented in \cite{multiresolution_multitask_gp}. 

\section{Availability of labelled data}\label{data_labels_sec}

As we have seen, the Bayesian paradigm provides a fertile ground for developing a rich variety of techniques relevant to transfer learning in both supervised \cite{pretrain_loss_deep}, \cite{power_prior_1}, \cite{ibrahim_2015}, \cite{opt_BTL}, \cite{opt_BTR}, \cite{opt_count}, \cite{pan_cancer_surv} and unsupervised settings \cite{de_vito_2019}, \cite{de_vito_2021}, \cite{chandra2023}, \cite{Sotiropoulos2013-cb}, \cite{Sotiropoulos2016-jj}. 
When the focus is on prediction, there are often challenges presented by the limited availability of labelled data. The term {\em semi-supervised} learning refers to the case in which labels are only available for a subset of the samples. The joint modeling approaches for overlapping variable transfer learning described in the previous section can trivially handle semi-supervised settings, as missing labels are just one type of missing data.

Particularly challenging are cases of {\em one-shot} and {\em few-shot} learning, which refers to having only a single or a few labelled samples in the target domain, respectively. For articles proposing Bayesian approaches to handle semi-supervised learning and these cases of a tiny number of labelled target data, refer to 
    \cite{bayesian_metalearning},
\cite{amortized_meta_few_shot},
\cite{hier_bayes_one_shot},
\cite{one_shot_character_2011}, \cite{one_shot_character_2013}, \cite{one_shot_character_2015}. 
In such cases, performance is critically dependent on borrowing of information from labelled data in the source domains. Common examples include classification based on images or audio (\cite{one_shot_character_2011}, \cite{one_shot_character_2013}, \cite{one_shot_character_2015}). For example, we may have many labelled examples of different individuals handwriting but only a tiny number for the individual of interest. 

In the transfer learning literature, \textit{inductive transfer learning} refers to the case where target domain labels are available, while \textit{transductive transfer learning} has labels available only in the source domains \cite{survey_transfer_learning}.
Of course, if there are certain systematic differences between the source and target domains, accurate transductive transfer may be impossible \cite{impossibility_ben_david}. Nonetheless, there is a rich PAC-Bayesian literature on this topic (\cite{germain2013}, \cite{germain2016}, \cite{germain2020}, \cite{pac_bayes_multiclass}) which specifies conditions allowing for successful training of the target learner in the absence of target labels in classification settings. They provide theoretical upper bounds on the expected error on the target domains of a Gibbs classifier depending on various measures of divergence between the distributions of predictors and labels of both domains as well as the properties of the set of voter algorithms from which the Gibbs classifier is constructed.

\section{Simulation illustration}\label{simulation_study}

In order to illustrate Bayesian transfer learning in practice, we run a simulation experiment focused on the problem of transfer learning targeting the covariance and/or precision matrix of a high-dimensional multivariate Gaussian distribution. In particular, there are data collected on the same set of variables for subjects in different groups and we would like to allow the covariance/precision to vary across groups, while borrowing information. This is a natural setting for both Bayesian multi-study factor analysis models and frequentist competitors focused on transfer learning in precision matrix estimation.

Specifically, we compare Bayesian Subspace Factor Analysis (SUFA) \cite{chandra2023} presented in section \ref{shared_latent} with the frequentist Trans-CLIME method proposed by Li \textit{et al.} \cite{trans_clime}. Trans-CLIME is a transfer learning extension of constrained
$L_1$ minimization for inverse matrix estimation (CLIME) \cite{clime}. We also include the frequentist estimator proposed by Guo \textit{et al.} \cite{mt-glasso}, which Li \textit{et al.} \cite{trans_clime} refer to as multitask graphical lasso (MT-Glasso). While \cite{chandra2023} focused on inferring the covariance, SUFA can just as easily be used to infer any functional of the covariance including the precision. We focus our comparisons on precision estimation, as this was the emphasis in 
\cite{trans_clime} and \cite{mt-glasso}.

We generate the synthetic data in two domains, $T$ (target) and $S$ (source), with sample sizes $n_T = 40$ and $n_S = 80$, respectively, and consider the data dimensions $p \in \{40, 60, 80, \dots, 280\}$. We generate 50 replicated datasets for each value of $p$. We report the average Frobenius and $L_1$ norm of the errors between the estimated and true values for the target precision matrix. For each value of $p$ considered, we generate the data using a fixed true precision matrix across all the simulation replicates. We present the results in Figure 3.

\begin{figure*}[]
\centering
    \includegraphics[scale=0.75]{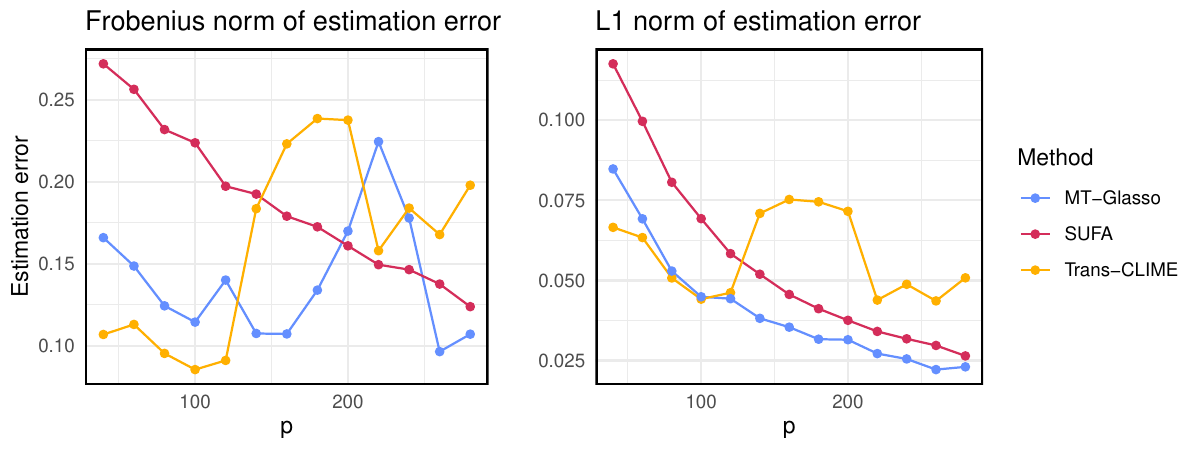}
      \caption{Frobenius and $L_1$ norm errors of target precision matrix estimation for SUFA, Trans-CLIME and MT-Glasso over varying dimension $p$.}
\end{figure*}

As we can see, in terms of the $L_1$ norm of the error, the performance tends to improve significantly with growing dimension for SUFA and MT-Glasso, and to an extent for Trans-CLIME as well. This somewhat counterintuitive phenomenon, known as the blessing of dimensionality \cite{cao2019} is commonly encountered in the covariance and precision estimation literature \cite{molstad2022}, \cite{xu2022proximal}, \cite{cao2019}, \cite{blessing_dim}. For the Frobenius norm error we see more instability in the performance of MT-Glasso and Trans-CLIME. Intuitively, the $L_1$ norm might be more favorable for these two methods, since they are both built on $L_1$ minimization. 

Since Trans-CLIME does not guarantee positive definiteness and invertibility of the estimated precision matrices, it is problematic to use it for covariance estimation; even after selecting only the synthetic datasets for which Trans-CLIME did produce invertible precision matrices, the resulting covariance estimates were highly unstable and overall significantly worse than those given by SUFA. MT-Glasso produces invertible precision matrices but 
also yielded covariance estimates further from the truth than SUFA estimates.

Hence, we find that this particular Bayesian approach to transfer learning based on a shared latent space model is competitive with frequentist counterparts even on tasks it was not built for (estimating the precision instead of the covariance). Indeed, as we have partially illustrated, since we have a posterior for the covariance that has support on the space of positive semidefinite covariances, we can provide Bayes estimates and posterior credible intervals providing uncertainty quantification for any functional of the covariance of interest. Hence, from a single Bayesian analysis, we can provide multiple results of interest that are all internally coherent.

\section{Discussion}\label{discussion}
Transfer learning is a timely problem  given the abundance of datasets from related domains. 
In many applications, there is simply not enough data from the domain of interest to support reliable inference and accurate predictions as we seek to fit increasingly complex models. Hence, it becomes critical to cleverly borrow information from available ``source'' datasets.

Choosing the appropriate strength and structure of information transfer between the domains remains one of the key challenges. The Bayesian paradigm offers a wide variety of approaches to transfer learning, including shared parameters, hierarchical and random effects models, shared latent space, and network transfer methods. There is a rich literature developing and applying these approaches in transfer learning settings, even though most often ``transfer learning'' is not mentioned in the associated papers.

This article has focused on providing a flavor for some of the interesting directions that are possible in terms of Bayesian transfer learning, but has not attempted a comprehensive overview of the massive relevant literature. Most of the transfer learning literature has focused on the simplest ``common variables" case,
 where data from different domains consist of the same variables measured across different subjects. Bayesian ideas applied to transfer learning can be particularly useful in the more challenging settings presented in Section \ref{interp_intra_sec}, where these existing methods largely do not apply. %

We have purposely focused much of our attention on shared latent space-type models for Bayesian transfer learning, ranging from multi-study factor analysis to multi-group Bayesian nonparametric models. We focused on these areas because the associated models are not as well known to the broad community but are very practically useful, including in challenging high-dimensional and complex structured data cases. In addition, there have been interesting recent developments that we have highlighted, while sketching out some promising directions for ongoing research. This includes extending Bayesian continuous latent factor modeling approaches to transfer learning settings. Our view is that more careful statistical models will tend to dominate over-parametrized black boxes, such as VAEs, in many settings.

An additional interesting area for future research is Bayesian transfer learning involving deep neural networks.
While in recent years there have been papers taking some early steps in this area \cite{pretrain_loss_deep}, \cite{multi_source_deep_tl}, there is plenty of potential for further impactful developments in this field, especially given the importance of transfer learning to deep neural networks training due to their data-hungry nature.    


\section{Acknowledgements}
This project has received funding from the European Research Council (ERC) under the European Union’s Horizon 2020 Research and Innovation Programme (grant agreement No. 856506),
United States National Institutes of Health (R01ES035625), Office of Naval Research (N00014-21-1-2510), National Institute of Environmental Health Sciences (R01ES027498), and National Science Foundation (DMS-2230074). Suder's contributions were supported in part by the Duke Summer Research Fellowship.

\bibliographystyle{imsart-number} 
\bibliography{references}       

\begin{thebibliography}{110}

\bibitem{rodriguez_2008}
\begin{barticle}[author]
\bauthor{\bsnm{Abel~Rodríguez},~\bfnm{David B~Dunson}\binits{D.~B.~D.}} \AND \bauthor{\bsnm{Gelfand},~\bfnm{Alan~E}\binits{A.~E.}}
(\byear{2008}).
\btitle{The nested Dirichlet process}.
\bjournal{Journal of the American Statistical Association}.
\end{barticle}
\endbibitem

\bibitem{gan_forgetting}
\begin{barticle}[author]
\bauthor{\bsnm{Avrahami},~\bfnm{Omri}\binits{O.}}, \bauthor{\bsnm{Lischinski},~\bfnm{Dani}\binits{D.}} \AND \bauthor{\bsnm{Fried},~\bfnm{Ohad}\binits{O.}}
(\byear{2021}).
\btitle{GAN cocktail: Mixing GANs without dataset access}.
\bjournal{European Conference on Computer Vision}.
\end{barticle}
\endbibitem

\bibitem{Lanczos}
\begin{barticle}[author]
\bauthor{\bsnm{Baglama},~\bfnm{James}\binits{J.}} \AND \bauthor{\bsnm{Reichel},~\bfnm{Lothar}\binits{L.}}
(\byear{2005}).
\btitle{Augmented implicitly restarted Lanczos bidiagonalization methods}.
\bjournal{SIAM Journal on Scientific Computing}.
\end{barticle}
\endbibitem

\bibitem{prot_network2}
\begin{barticle}[author]
\bauthor{\bsnm{Baladandayuthapani},~\bfnm{Veerabhadran}\binits{V.}}, \bauthor{\bsnm{Talluri},~\bfnm{Rajesh}\binits{R.}}, \bauthor{\bsnm{Ji},~\bfnm{Yuan}\binits{Y.}}, \bauthor{\bsnm{Coombes},~\bfnm{Kevin}\binits{K.}}, \bauthor{\bsnm{Lu},~\bfnm{Yiling}\binits{Y.}}, \bauthor{\bsnm{Hennessy},~\bfnm{Bryan}\binits{B.}}, \bauthor{\bsnm{Davies},~\bfnm{Michael}\binits{M.}} \AND \bauthor{\bsnm{Mallick},~\bfnm{Bani}\binits{B.}}
(\byear{2014}).
\btitle{Bayesian sparse graphical models for classification with application to protein expression data}.
\bjournal{The Annals of Applied Statistics}.
\end{barticle}
\endbibitem

\bibitem{hier_GP_2005}
\begin{barticle}[author]
\bauthor{\bsnm{Behseta},~\bfnm{Sam}\binits{S.}}, \bauthor{\bsnm{Kass},~\bfnm{Robert~E}\binits{R.~E.}} \AND \bauthor{\bsnm{Wallstrom},~\bfnm{Garrick~L}\binits{G.~L.}}
(\byear{2005}).
\btitle{Hierarchical models for assessing variability among functions}.
\bjournal{Biometrika}.
\end{barticle}
\endbibitem

\bibitem{DL_prior}
\begin{barticle}[author]
\bauthor{\bsnm{Bhattacharya},~\bfnm{Anirban}\binits{A.}}, \bauthor{\bsnm{Pati},~\bfnm{Debdeep}\binits{D.}}, \bauthor{\bsnm{Pillai},~\bfnm{Natesh~S.}\binits{N.~S.}} \AND \bauthor{\bsnm{Dunson},~\bfnm{David~B.}\binits{D.~B.}}
(\byear{2015}).
\btitle{Dirichlet–Laplace priors for optimal shrinkage}.
\bjournal{Journal of the American Statistical Association}.
\end{barticle}
\endbibitem

\bibitem{cancer_rna}
\begin{barticle}[author]
\bauthor{\bsnm{Boluki},~\bfnm{Shahin}\binits{S.}}, \bauthor{\bsnm{Qian},~\bfnm{Xiaoning}\binits{X.}} \AND \bauthor{\bsnm{Dougherty},~\bfnm{Edward~R}\binits{E.~R.}}
(\byear{2021}).
\btitle{{Optimal Bayesian supervised domain adaptation for RNA sequencing data}}.
\bjournal{Bioinformatics}.
\end{barticle}
\endbibitem

\bibitem{deep_forgetting}
\begin{barticle}[author]
\bauthor{\bsnm{Boschini},~\bfnm{Matteo}\binits{M.}}, \bauthor{\bsnm{Bonicelli},~\bfnm{Lorenzo}\binits{L.}}, \bauthor{\bsnm{Porrello},~\bfnm{Angelo}\binits{A.}}, \bauthor{\bsnm{Bellitto},~\bfnm{Giovanni}\binits{G.}}, \bauthor{\bsnm{Pennisi},~\bfnm{Matteo}\binits{M.}}, \bauthor{\bsnm{Palazzo},~\bfnm{Simone}\binits{S.}}, \bauthor{\bsnm{Spampinato},~\bfnm{Concetto}\binits{C.}} \AND \bauthor{\bsnm{Calderara},~\bfnm{Simone}\binits{S.}}
(\byear{2022}).
\btitle{Transfer without forgetting}.
\bjournal{European Conference on Computer Vision}.
\end{barticle}
\endbibitem

\bibitem{bu2023inferring}
\begin{barticle}[author]
\bauthor{\bsnm{Bu},~\bfnm{Fan}\binits{F.}}, \bauthor{\bsnm{Kagaayi},~\bfnm{Joseph}\binits{J.}}, \bauthor{\bsnm{Grabowski},~\bfnm{Kate}\binits{K.}}, \bauthor{\bsnm{Ratmann},~\bfnm{Oliver}\binits{O.}} \AND \bauthor{\bsnm{Xu},~\bfnm{Jason}\binits{J.}}
(\byear{2023}).
\btitle{Inferring HIV transmission patterns from viral deep-sequence data via latent typed point processes. \textit{arXiv preprint \href{https://arxiv.org/pdf/2302.11567.pdf}{arXiv 2302.11567}}}.
\end{barticle}
\endbibitem

\bibitem{hier_proc}
\begin{barticle}[author]
\bauthor{\bsnm{Camerlenghi},~\bfnm{Federico}\binits{F.}}, \bauthor{\bsnm{Lijoi},~\bfnm{Antonio}\binits{A.}}, \bauthor{\bsnm{Orbanz},~\bfnm{Peter}\binits{P.}} \AND \bauthor{\bsnm{Pr{\"u}nster},~\bfnm{Igor}\binits{I.}}
(\byear{2019}).
\btitle{{Distribution theory for hierarchical processes}}.
\bjournal{The Annals of Statistics}.
\end{barticle}
\endbibitem

\bibitem{Carvalho2008}
\begin{barticle}[author]
\bauthor{\bsnm{Carvalho},~\bfnm{Carlos~M}\binits{C.~M.}}, \bauthor{\bsnm{Chang},~\bfnm{Jeffrey}\binits{J.}}, \bauthor{\bsnm{Lucas},~\bfnm{Joseph~E}\binits{J.~E.}}, \bauthor{\bsnm{Nevins},~\bfnm{Joseph~R}\binits{J.~R.}}, \bauthor{\bsnm{Wang},~\bfnm{Quanli}\binits{Q.}} \AND \bauthor{\bsnm{West},~\bfnm{Mike}\binits{M.}}
(\byear{2008}).
\btitle{High-dimensional sparse factor modeling: Applications in gene expression genomics}.
\bjournal{Journal of the American Statistical Association}.
\end{barticle}
\endbibitem

\bibitem{auto_vae}
\begin{barticle}[author]
\bauthor{\bsnm{Cemgil},~\bfnm{Taylan}\binits{T.}}, \bauthor{\bsnm{Ghaisas},~\bfnm{Sumedh}\binits{S.}}, \bauthor{\bsnm{Dvijotham},~\bfnm{Krishnamurthy}\binits{K.}}, \bauthor{\bsnm{Gowal},~\bfnm{Sven}\binits{S.}} \AND \bauthor{\bsnm{Kohli},~\bfnm{Pushmeet}\binits{P.}}
(\byear{2020}).
\btitle{The autoencoding variational autoencoder}.
\bjournal{Advances in Neural Information Processing Systems}.
\end{barticle}
\endbibitem

\bibitem{chakrabarti2023graphical}
\begin{bmisc}[author]
\bauthor{\bsnm{Chakrabarti},~\bfnm{Arhit}\binits{A.}}, \bauthor{\bsnm{Ni},~\bfnm{Yang}\binits{Y.}}, \bauthor{\bsnm{Morris},~\bfnm{Ellen Ruth~A.}\binits{E.~R.~A.}}, \bauthor{\bsnm{Salinas},~\bfnm{Michael~L.}\binits{M.~L.}}, \bauthor{\bsnm{Chapkin},~\bfnm{Robert~S.}\binits{R.~S.}} \AND \bauthor{\bsnm{Mallick},~\bfnm{Bani~K.}\binits{B.~K.}}
(\byear{2023}).
\btitle{Graphical Dirichlet process for clustering non-exchangeable grouped data. \textit{arXiv preprint \href{https://arxiv.org/pdf/2302.09111.pdf}{arXiv 2302.09111}}}.
\end{bmisc}
\endbibitem

\bibitem{Clustering_Chandra}
\begin{barticle}[author]
\bauthor{\bsnm{Chandra},~\bfnm{Noirrit~Kiran}\binits{N.~K.}}, \bauthor{\bsnm{Canale},~\bfnm{Antonio}\binits{A.}} \AND \bauthor{\bsnm{Dunson},~\bfnm{David~B.}\binits{D.~B.}}
(\byear{2023}).
\btitle{Escaping the curse of dimensionality in Bayesian model-based clustering}.
\bjournal{Journal of Machine Learning Research}.
\end{barticle}
\endbibitem

\bibitem{chandra2023}
\begin{barticle}[author]
\bauthor{\bsnm{Chandra},~\bfnm{Noirrit~Kiran}\binits{N.~K.}}, \bauthor{\bsnm{Dunson},~\bfnm{David~B.}\binits{D.~B.}} \AND \bauthor{\bsnm{Xu},~\bfnm{Jason}\binits{J.}}
(\byear{2023}).
\btitle{Inferring covariance structure from multiple data sources via subspace factor analysis. \textit{arXiv preprint \href{https://arxiv.org/pdf/2305.04113.pdf}{arXiv 2305.04113}}}.
\end{barticle}
\endbibitem

\bibitem{multi_source_deep_tl}
\begin{barticle}[author]
\bauthor{\bsnm{Chandra},~\bfnm{Rohitash}\binits{R.}} \AND \bauthor{\bsnm{Kapoor},~\bfnm{Arpit}\binits{A.}}
(\byear{2020}).
\btitle{Bayesian neural multi-source transfer learning}.
\bjournal{Neurocomputing}.
\end{barticle}
\endbibitem

\bibitem{power_prior_hierarchical_models}
\begin{barticle}[author]
\bauthor{\bsnm{Chen},~\bfnm{Ming-Hui}\binits{M.-H.}} \AND \bauthor{\bsnm{Ibrahim},~\bfnm{Joseph}\binits{J.}}
(\byear{2006}).
\btitle{The relationship between the power prior and hierarchical models}.
\bjournal{Bayesian Analysis}.
\end{barticle}
\endbibitem

\bibitem{power_prior_1}
\begin{barticle}[author]
\bauthor{\bsnm{Chen},~\bfnm{Ming-Hui}\binits{M.-H.}} \AND \bauthor{\bsnm{Ibrahim},~\bfnm{Joseph~G.}\binits{J.~G.}}
(\byear{2000}).
\btitle{{Power prior distributions for regression models}}.
\bjournal{Statistical Science}.
\end{barticle}
\endbibitem

\bibitem{partial_borrowing_power_prior}
\begin{barticle}[author]
\bauthor{\bsnm{Chen},~\bfnm{Ming-Hui}\binits{M.-H.}}, \bauthor{\bsnm{Ibrahim},~\bfnm{Joseph~G.}\binits{J.~G.}}, \bauthor{\bsnm{Lam},~\bfnm{Peter}\binits{P.}}, \bauthor{\bsnm{Yu},~\bfnm{Alan}\binits{A.}} \AND \bauthor{\bsnm{Zhang},~\bfnm{Yuanye}\binits{Y.}}
(\byear{2011}).
\btitle{Bayesian design of noninferiority trials for medical devices using historical data}.
\bjournal{Biometrics}.
\end{barticle}
\endbibitem

\bibitem{post_collapse}
\begin{barticle}[author]
\bauthor{\bsnm{Dai},~\bfnm{Bin}\binits{B.}}, \bauthor{\bsnm{Wang},~\bfnm{Ziyu}\binits{Z.}} \AND \bauthor{\bsnm{Wipf},~\bfnm{David}\binits{D.}}
(\byear{2020}).
\btitle{The usual suspects? Reassessing blame for VAE posterior collapse}.
\bjournal{International Conference on Machine Learning}.
\end{barticle}
\endbibitem

\bibitem{hier_GP_2019}
\begin{barticle}[author]
\bauthor{\bsnm{Daniel R.~Kowal},~\bfnm{David S.~Matteson}\binits{D.~S.~M.}} \AND \bauthor{\bsnm{Ruppert},~\bfnm{David}\binits{D.}}
(\byear{2019}).
\btitle{Functional autoregression for sparsely sampled data}.
\bjournal{Journal of Business \& Economic Statistics}.
\end{barticle}
\endbibitem

\bibitem{impossibility_ben_david}
\begin{barticle}[author]
\bauthor{\bsnm{David},~\bfnm{Shai~Ben}\binits{S.~B.}}, \bauthor{\bsnm{Lu},~\bfnm{Tyler}\binits{T.}}, \bauthor{\bsnm{Luu},~\bfnm{Teresa}\binits{T.}} \AND \bauthor{\bsnm{Pal},~\bfnm{David}\binits{D.}}
(\byear{2010}).
\btitle{Impossibility theorems for domain adaptation}.
\bjournal{International Conference on Artificial Intelligence and Statistics}.
\end{barticle}
\endbibitem

\bibitem{de_vito_2019}
\begin{barticle}[author]
\bauthor{\bsnm{De~Vito},~\bfnm{Roberta}\binits{R.}}, \bauthor{\bsnm{Bellio},~\bfnm{Ruggero}\binits{R.}}, \bauthor{\bsnm{Trippa},~\bfnm{Lorenzo}\binits{L.}} \AND \bauthor{\bsnm{Parmigiani},~\bfnm{Giovanni}\binits{G.}}
(\byear{2019}).
\btitle{Multi-study factor analysis}.
\bjournal{Biometrics}.
\end{barticle}
\endbibitem

\bibitem{imagenet}
\begin{barticle}[author]
\bauthor{\bsnm{Deng},~\bfnm{Jia}\binits{J.}}, \bauthor{\bsnm{Dong},~\bfnm{Wei}\binits{W.}}, \bauthor{\bsnm{Socher},~\bfnm{Richard}\binits{R.}}, \bauthor{\bsnm{Li},~\bfnm{Li-Jia}\binits{L.-J.}}, \bauthor{\bsnm{Li},~\bfnm{Kai}\binits{K.}} \AND \bauthor{\bsnm{Fei-Fei},~\bfnm{Li}\binits{L.}}
(\byear{2009}).
\btitle{ImageNet: A large-scale hierarchical image database}.
\bjournal{IEEE Conference on Computer Vision and Pattern Recognition}.
\end{barticle}
\endbibitem

\bibitem{network_meta_Dias2011}
\begin{bbook}[author]
\bauthor{\bsnm{Dias},~\bfnm{Sofia}\binits{S.}}, \bauthor{\bsnm{Welton},~\bfnm{Nicky~J}\binits{N.~J.}}, \bauthor{\bsnm{Sutton},~\bfnm{Alex~J}\binits{A.~J.}} \AND \bauthor{\bsnm{Ades},~\bfnm{A~E}\binits{A.~E.}}
(\byear{2014}).
\btitle{{NICE} {DSU} technical support document 2: A generalised linear modelling framework for pairwise and network {meta-analysis} of randomised controlled trials}.
\bpublisher{National Institute for Health and Care Excellence (NICE)}.
\end{bbook}
\endbibitem

\bibitem{tibshiriani_cooperative_learn}
\begin{barticle}[author]
\bauthor{\bsnm{Ding},~\bfnm{Daisy~Yi}\binits{D.~Y.}}, \bauthor{\bsnm{Li},~\bfnm{Shuangning}\binits{S.}}, \bauthor{\bsnm{Narasimhan},~\bfnm{Balasubramanian}\binits{B.}} \AND \bauthor{\bsnm{Tibshirani},~\bfnm{Robert}\binits{R.}}
(\byear{2022}).
\btitle{Cooperative learning for multiview analysis}.
\bjournal{Proceedings of the National Academy of Sciences}.
\end{barticle}
\endbibitem

\bibitem{normalized_power_prior}
\begin{barticle}[author]
\bauthor{\bsnm{Duan},~\bfnm{Yuyan}\binits{Y.}}, \bauthor{\bsnm{Ye},~\bfnm{Keying}\binits{K.}} \AND \bauthor{\bsnm{Smith},~\bfnm{Eric~P.}\binits{E.~P.}}
(\byear{2006}).
\btitle{Evaluating water quality using power priors to incorporate historical information}.
\bjournal{Environmetrics}.
\end{barticle}
\endbibitem

\bibitem{networks_latent}
\begin{barticle}[author]
\bauthor{\bsnm{Durante},~\bfnm{Daniele}\binits{D.}} \AND \bauthor{\bsnm{Dunson},~\bfnm{David~B.}\binits{D.~B.}}
(\byear{2018}).
\btitle{{Bayesian inference and testing of group differences in brain networks}}.
\bjournal{Bayesian Analysis}.
\end{barticle}
\endbibitem

\bibitem{continual_learning_4}
\begin{barticle}[author]
\bauthor{\bsnm{Ebrahimi},~\bfnm{Sayna}\binits{S.}}, \bauthor{\bsnm{Elhoseiny},~\bfnm{Mohamed}\binits{M.}}, \bauthor{\bsnm{Darrell},~\bfnm{Trevor}\binits{T.}} \AND \bauthor{\bsnm{Rohrbach},~\bfnm{Marcus}\binits{M.}}
(\byear{2020}).
\btitle{Uncertainty-guided continual learning with Bayesian neural networks}.
\bjournal{International Conference on Learning Representations}.
\end{barticle}
\endbibitem

\bibitem{multiview_GP_LVM}
\begin{barticle}[author]
\bauthor{\bsnm{Eleftheriadis},~\bfnm{Stefanos}\binits{S.}}, \bauthor{\bsnm{Rudovic},~\bfnm{Ognjen}\binits{O.}} \AND \bauthor{\bsnm{Pantic},~\bfnm{Maja}\binits{M.}}
(\byear{2014}).
\btitle{Discriminative shared gaussian processes for multiview and view-invariant facial expression recognition}.
\bjournal{IEEE Transactions on Image Processing}.
\end{barticle}
\endbibitem

\bibitem{ferrari2021}
\begin{barticle}[author]
\bauthor{\bsnm{Ferrari},~\bfnm{Federico}\binits{F.}} \AND \bauthor{\bsnm{Dunson},~\bfnm{David~B.}\binits{D.~B.}}
(\byear{2021}).
\btitle{Bayesian factor analysis for inference on interactions}.
\bjournal{Journal of the American Statistical Association}.
\end{barticle}
\endbibitem

\bibitem{germain2013}
\begin{barticle}[author]
\bauthor{\bsnm{Germain},~\bfnm{Pascal}\binits{P.}}, \bauthor{\bsnm{Habrard},~\bfnm{Amaury}\binits{A.}}, \bauthor{\bsnm{Laviolette},~\bfnm{François}\binits{F.}} \AND \bauthor{\bsnm{Morvant},~\bfnm{Emilie}\binits{E.}}
(\byear{2013}).
\btitle{A \uppercase{PAC-B}ayesian approach for domain adaptation with specialization to linear classifiers}.
\bjournal{International Conference on Machine Learning}.
\end{barticle}
\endbibitem

\bibitem{germain2016}
\begin{barticle}[author]
\bauthor{\bsnm{Germain},~\bfnm{Pascal}\binits{P.}}, \bauthor{\bsnm{Habrard},~\bfnm{Amaury}\binits{A.}}, \bauthor{\bsnm{Laviolette},~\bfnm{François}\binits{F.}} \AND \bauthor{\bsnm{Morvant},~\bfnm{Emilie}\binits{E.}}
(\byear{2016}).
\btitle{A new \uppercase{PAC-B}ayesian perspective on domain adaptation}.
\bjournal{International Conference on Machine Learning}.
\end{barticle}
\endbibitem

\bibitem{germain2020}
\begin{barticle}[author]
\bauthor{\bsnm{Germain},~\bfnm{Pascal}\binits{P.}}, \bauthor{\bsnm{Habrard},~\bfnm{Amaury}\binits{A.}}, \bauthor{\bsnm{Laviolette},~\bfnm{François}\binits{F.}} \AND \bauthor{\bsnm{Morvant},~\bfnm{Emilie}\binits{E.}}
(\byear{2020}).
\btitle{\uppercase{PAC-B}ayes and domain adaptation}.
\bjournal{Neurocomputing}.
\end{barticle}
\endbibitem

\bibitem{green_1995}
\begin{barticle}[author]
\bauthor{\bsnm{Green},~\bfnm{Peter~J.}\binits{P.~J.}}
(\byear{1995}).
\btitle{Reversible jump Markov chain Monte Carlo computation and Bayesian model determination}.
\bjournal{Biometrika}.
\end{barticle}
\endbibitem

\bibitem{mt-glasso}
\begin{barticle}[author]
\bauthor{\bsnm{Guo},~\bfnm{Jian}\binits{J.}}, \bauthor{\bsnm{Levina},~\bfnm{Elizaveta}\binits{E.}}, \bauthor{\bsnm{Michailidis},~\bfnm{George}\binits{G.}} \AND \bauthor{\bsnm{Zhu},~\bfnm{Ji}\binits{J.}}
(\byear{2011}).
\btitle{{Joint estimation of multiple graphical models}}.
\bjournal{Biometrika}.
\end{barticle}
\endbibitem

\bibitem{kernel_bayes_tl}
\begin{barticle}[author]
\bauthor{\bsnm{Gönen},~\bfnm{Mehmet}\binits{M.}} \AND \bauthor{\bsnm{Margolin},~\bfnm{A.~A.}\binits{A.~A.}}
(\byear{2014}).
\btitle{Kernelized Bayesian transfer learning}.
\bjournal{AAAI Conference on Artificial Intelligence}.
\end{barticle}
\endbibitem

\bibitem{osti_10096293}
\begin{barticle}[author]
\bauthor{\bsnm{Hajiramezanali},~\bfnm{Ehsan}\binits{E.}}, \bauthor{\bsnm{Zamani~Dadaneh},~\bfnm{Siamak}\binits{S.}}, \bauthor{\bsnm{Karbalayghareh},~\bfnm{Alireza}\binits{A.}}, \bauthor{\bsnm{Zhou},~\bfnm{Mingyuan}\binits{M.}} \AND \bauthor{\bsnm{Qian},~\bfnm{Xiaoning}\binits{X.}}
(\byear{2018}).
\btitle{Bayesian multi-domain learning for cancer subtype discovery from next-generation sequencing count data}.
\bjournal{Advances in Neural Information Processing Systems}.
\end{barticle}
\endbibitem

\bibitem{multiresolution_multitask_gp}
\begin{barticle}[author]
\bauthor{\bsnm{Hamelijnck},~\bfnm{Oliver}\binits{O.}}, \bauthor{\bsnm{Damoulas},~\bfnm{Theodoros}\binits{T.}}, \bauthor{\bsnm{Wang},~\bfnm{Kangrui}\binits{K.}} \AND \bauthor{\bsnm{Girolami},~\bfnm{Mark}\binits{M.}}
(\byear{2019}).
\btitle{Multi-resolution multi-task Gaussian processes}.
\bjournal{Advances in Neural Information Processing Systems}.
\end{barticle}
\endbibitem

\bibitem{hoff_2008}
\begin{barticle}[author]
\bauthor{\bsnm{Hoff},~\bfnm{Peter}\binits{P.}}
(\byear{2007}).
\btitle{Modeling homophily and stochastic equivalence in symmetric relational data}.
\bjournal{Advances in Neural Information Processing Systems}.
\end{barticle}
\endbibitem

\bibitem{metalearning_survey}
\begin{barticle}[author]
\bauthor{\bsnm{Hospedales},~\bfnm{T.}\binits{T.}}, \bauthor{\bsnm{Antoniou},~\bfnm{A.}\binits{A.}}, \bauthor{\bsnm{Micaelli},~\bfnm{P.}\binits{P.}} \AND \bauthor{\bsnm{Storkey},~\bfnm{A.}\binits{A.}}
(\byear{2022}).
\btitle{Meta-learning in neural networks: A survey}.
\bjournal{IEEE Transactions on Pattern Analysis and Machine Intelligence}.
\end{barticle}
\endbibitem

\bibitem{a0_criterion_1}
\begin{barticle}[author]
\bauthor{\bsnm{Ibrahim},~\bfnm{J.~G.}\binits{J.~G.}}, \bauthor{\bsnm{Chen},~\bfnm{M.~H.}\binits{M.~H.}} \AND \bauthor{\bsnm{Sinha},~\bfnm{D.}\binits{D.}}
(\byear{2001}).
\btitle{Bayesian semiparametric models for survival data with a cure fraction}.
\bjournal{Biometrics}.
\end{barticle}
\endbibitem

\bibitem{ibrahim_2015}
\begin{barticle}[author]
\bauthor{\bsnm{Ibrahim},~\bfnm{Joseph~G.}\binits{J.~G.}}, \bauthor{\bsnm{Chen},~\bfnm{Ming-Hui}\binits{M.-H.}}, \bauthor{\bsnm{Gwon},~\bfnm{Yeongjin}\binits{Y.}} \AND \bauthor{\bsnm{Chen},~\bfnm{Fang}\binits{F.}}
(\byear{2015}).
\btitle{The power prior: theory and applications}.
\bjournal{Statistics in Medicine}.
\end{barticle}
\endbibitem

\bibitem{optimality_power_prior}
\begin{barticle}[author]
\bauthor{\bsnm{Ibrahim},~\bfnm{Joseph~G}\binits{J.~G.}}, \bauthor{\bsnm{Chen},~\bfnm{Ming-Hui}\binits{M.-H.}} \AND \bauthor{\bsnm{Sinha},~\bfnm{Debajyoti}\binits{D.}}
(\byear{2003}).
\btitle{On optimality properties of the power prior}.
\bjournal{Journal of the American Statistical Association}.
\end{barticle}
\endbibitem

\bibitem{domain_inv_vae}
\begin{barticle}[author]
\bauthor{\bsnm{Ilse},~\bfnm{Maximilian}\binits{M.}}, \bauthor{\bsnm{Tomczak},~\bfnm{Jakub~M.}\binits{J.~M.}}, \bauthor{\bsnm{Louizos},~\bfnm{Christos}\binits{C.}} \AND \bauthor{\bsnm{Welling},~\bfnm{Max}\binits{M.}}
(\byear{2020}).
\btitle{DIVA: Domain invariant variational autoencoders}.
\bjournal{Conference on Medical Imaging with Deep Learning}.
\end{barticle}
\endbibitem

\bibitem{continual_learning_1}
\begin{barticle}[author]
\bauthor{\bsnm{Kapoor},~\bfnm{Sanyam}\binits{S.}}, \bauthor{\bsnm{Karaletsos},~\bfnm{Theofanis}\binits{T.}} \AND \bauthor{\bsnm{Bui},~\bfnm{Thang~D}\binits{T.~D.}}
(\byear{2021}).
\btitle{Variational auto-regressive Gaussian processes for continual learning}.
\bjournal{International Conference on Machine Learning}.
\end{barticle}
\endbibitem

\bibitem{opt_BTL}
\begin{barticle}[author]
\bauthor{\bsnm{Karbalayghareh},~\bfnm{Alireza}\binits{A.}}, \bauthor{\bsnm{Qian},~\bfnm{Xiaoning}\binits{X.}} \AND \bauthor{\bsnm{Dougherty},~\bfnm{Edward~R.}\binits{E.~R.}}
(\byear{2018}).
\btitle{Optimal Bayesian transfer learning}.
\bjournal{IEEE Transactions on Signal Processing}.
\end{barticle}
\endbibitem

\bibitem{opt_BTR}
\begin{barticle}[author]
\bauthor{\bsnm{Karbalayghareh},~\bfnm{Alireza}\binits{A.}}, \bauthor{\bsnm{Qian},~\bfnm{Xiaoning}\binits{X.}} \AND \bauthor{\bsnm{Dougherty},~\bfnm{Edward~R.}\binits{E.~R.}}
(\byear{2018}).
\btitle{Optimal Bayesian transfer regression}.
\bjournal{IEEE Signal Processing Letters}.
\end{barticle}
\endbibitem

\bibitem{opt_count}
\begin{barticle}[author]
\bauthor{\bsnm{Karbalayghareh},~\bfnm{Alireza}\binits{A.}}, \bauthor{\bsnm{Qian},~\bfnm{Xiaoning}\binits{X.}} \AND \bauthor{\bsnm{Dougherty},~\bfnm{Edward~R.}\binits{E.~R.}}
(\byear{2021}).
\btitle{Optimal Bayesian transfer learning for count data}.
\bjournal{IEEE/ACM Transactions on Computational Biology and Bioinformatics}.
\end{barticle}
\endbibitem

\bibitem{domain_adapt_tl}
\begin{barticle}[author]
\bauthor{\bsnm{Kouw},~\bfnm{Wouter~M.}\binits{W.~M.}} \AND \bauthor{\bsnm{Loog},~\bfnm{Marco}\binits{M.}}
(\byear{2019}).
\btitle{An introduction to domain adaptation and transfer learning. \textit{arXiv preprint \href{https://arxiv.org/pdf/1812.11806.pdf}{arXiv 1812.11806}}}.
\end{barticle}
\endbibitem

\bibitem{continual_learning_3}
\begin{barticle}[author]
\bauthor{\bsnm{Kumar},~\bfnm{Abhishek}\binits{A.}}, \bauthor{\bsnm{Chatterjee},~\bfnm{Sunabha}\binits{S.}} \AND \bauthor{\bsnm{Rai},~\bfnm{Piyush}\binits{P.}}
(\byear{2021}).
\btitle{Bayesian structural adaptation for continual learning}.
\bjournal{International Conference on Machine Learning}.
\end{barticle}
\endbibitem

\bibitem{open_images_v4}
\begin{barticle}[author]
\bauthor{\bsnm{Kuznetsova},~\bfnm{Alina}\binits{A.}}, \bauthor{\bsnm{Rom},~\bfnm{Hassan}\binits{H.}}, \bauthor{\bsnm{Alldrin},~\bfnm{Neil}\binits{N.}}, \bauthor{\bsnm{Uijlings},~\bfnm{Jasper}\binits{J.}}, \bauthor{\bsnm{Krasin},~\bfnm{Ivan}\binits{I.}}, \bauthor{\bsnm{Pont-Tuset},~\bfnm{Jordi}\binits{J.}}, \bauthor{\bsnm{Kamali},~\bfnm{Shahab}\binits{S.}}, \bauthor{\bsnm{Popov},~\bfnm{Stefan}\binits{S.}}, \bauthor{\bsnm{Malloci},~\bfnm{Matteo}\binits{M.}}, \bauthor{\bsnm{Kolesnikov},~\bfnm{Alexander}\binits{A.}}, \bauthor{\bsnm{Duerig},~\bfnm{Tom}\binits{T.}} \AND \bauthor{\bsnm{Ferrari},~\bfnm{Vittorio}\binits{V.}}
(\byear{2020}).
\btitle{The open images dataset V4}.
\bjournal{International Journal of Computer Vision}.
\end{barticle}
\endbibitem

\bibitem{one_shot_character_2011}
\begin{barticle}[author]
\bauthor{\bsnm{Lake},~\bfnm{Brenden~M.}\binits{B.~M.}}, \bauthor{\bsnm{Salakhutdinov},~\bfnm{Ruslan}\binits{R.}}, \bauthor{\bsnm{Gross},~\bfnm{Jason}\binits{J.}} \AND \bauthor{\bsnm{Tenenbaum},~\bfnm{Joshua~B.}\binits{J.~B.}}
(\byear{2011}).
\btitle{One shot learning of simple visual concepts}.
\bjournal{Cognitive Science}.
\end{barticle}
\endbibitem

\bibitem{one_shot_character_2015}
\begin{barticle}[author]
\bauthor{\bsnm{Lake},~\bfnm{Brenden~M.}\binits{B.~M.}}, \bauthor{\bsnm{Salakhutdinov},~\bfnm{Ruslan}\binits{R.}} \AND \bauthor{\bsnm{Tenenbaum},~\bfnm{Joshua~B.}\binits{J.~B.}}
(\byear{2015}).
\btitle{Human-level concept learning through probabilistic program induction}.
\bjournal{Science}.
\end{barticle}
\endbibitem

\bibitem{one_shot_character_2013}
\begin{barticle}[author]
\bauthor{\bsnm{Lake},~\bfnm{Brenden~M}\binits{B.~M.}}, \bauthor{\bsnm{Salakhutdinov},~\bfnm{Russ~R}\binits{R.~R.}} \AND \bauthor{\bsnm{Tenenbaum},~\bfnm{Josh}\binits{J.}}
(\byear{2013}).
\btitle{One-shot learning by inverting a compositional causal process}.
\bjournal{Advances in Neural Information Processing Systems}.
\end{barticle}
\endbibitem

\bibitem{hier_GP_LVM}
\begin{barticle}[author]
\bauthor{\bsnm{Lawrence},~\bfnm{Neil~D.}\binits{N.~D.}} \AND \bauthor{\bsnm{Moore},~\bfnm{Andrew~J.}\binits{A.~J.}}
(\byear{2007}).
\btitle{Hierarchical Gaussian process latent variable models}.
\bjournal{International Conference on Machine Learning}.
\end{barticle}
\endbibitem

\bibitem{leblanc2023}
\begin{bmisc}[author]
\bauthor{\bsnm{LeBlanc},~\bfnm{Patrick~M.}\binits{P.~M.}} \AND \bauthor{\bsnm{Banks},~\bfnm{David}\binits{D.}}
(\byear{2023}).
\btitle{Time-varying Bayesian network meta-analysis. \textit{arXiv preprint \href{https://arxiv.org/pdf/2211.08312.pdf}{arXiv 2211.08312}}}.
\end{bmisc}
\endbibitem

\bibitem{lock_dunson_shared_kernel}
\begin{barticle}[author]
\bauthor{\bsnm{Lock},~\bfnm{Eric~F.}\binits{E.~F.}} \AND \bauthor{\bsnm{Dunson},~\bfnm{David~B.}\binits{D.~B.}}
(\byear{2015}).
\btitle{Shared kernel Bayesian screening}.
\bjournal{Biometrika}.
\end{barticle}
\endbibitem

\bibitem{deep_tl_long}
\begin{barticle}[author]
\bauthor{\bsnm{Long},~\bfnm{Mingsheng}\binits{M.}}, \bauthor{\bsnm{Cao},~\bfnm{Yue}\binits{Y.}}, \bauthor{\bsnm{Wang},~\bfnm{Jianmin}\binits{J.}} \AND \bauthor{\bsnm{Jordan},~\bfnm{Michael~I.}\binits{M.~I.}}
(\byear{2015}).
\btitle{Learning Transferable Features with Deep Adaptation Networks}.
\bjournal{International Conference on Machine Learning}.
\end{barticle}
\endbibitem

\bibitem{west_factor_models}
\begin{barticle}[author]
\bauthor{\bsnm{Lopes},~\bfnm{Hedibert~Freitas}\binits{H.~F.}} \AND \bauthor{\bsnm{West},~\bfnm{Mike}\binits{M.}}
(\byear{2004}).
\btitle{Bayesian model assessment in factor analysis}.
\bjournal{Statistica Sinica}.
\end{barticle}
\endbibitem

\bibitem{network_meta_2004}
\begin{barticle}[author]
\bauthor{\bsnm{Lu},~\bfnm{G.}\binits{G.}} \AND \bauthor{\bsnm{Ades},~\bfnm{A.~E.}\binits{A.~E.}}
(\byear{2004}).
\btitle{Combination of direct and indirect evidence in mixed treatment comparisons}.
\bjournal{Statistics in Medicine}.
\end{barticle}
\endbibitem

\bibitem{bnma_2006}
\begin{barticle}[author]
\bauthor{\bsnm{Lu},~\bfnm{Guobing}\binits{G.}} \AND \bauthor{\bsnm{Ades},~\bfnm{A.~E.}\binits{A.~E.}}
(\byear{2006}).
\btitle{Assessing evidence inconsistency in mixed treatment comparisons}.
\bjournal{Journal of the American Statistical Association}.
\end{barticle}
\endbibitem

\bibitem{deep_tl_luo}
\begin{barticle}[author]
\bauthor{\bsnm{Luo},~\bfnm{Zelun}\binits{Z.}}, \bauthor{\bsnm{Zou},~\bfnm{Yuliang}\binits{Y.}}, \bauthor{\bsnm{Hoffman},~\bfnm{Judy}\binits{J.}} \AND \bauthor{\bsnm{Fei-Fei},~\bfnm{Li~F}\binits{L.~F.}}
(\byear{2017}).
\btitle{Label Efficient Learning of Transferable Representations across Domains and Tasks}.
\bjournal{Advances in Neural Information Processing Systems}.
\end{barticle}
\endbibitem

\bibitem{pan_cancer_surv}
\begin{barticle}[author]
\bauthor{\bsnm{Maity},~\bfnm{Arnab}\binits{A.}}, \bauthor{\bsnm{Bhattacharya},~\bfnm{Anirban}\binits{A.}}, \bauthor{\bsnm{Mallick},~\bfnm{Bani}\binits{B.}} \AND \bauthor{\bsnm{Baladandayuthapani},~\bfnm{Veerabhadran}\binits{V.}}
(\byear{2019}).
\btitle{Bayesian data integration and variable selection for pan‐cancer survival prediction using protein expression data}.
\bjournal{Biometrics}.
\end{barticle}
\endbibitem

\bibitem{cat_forgetting}
\begin{bincollection}[author]
\bauthor{\bsnm{McCloskey},~\bfnm{Michael}\binits{M.}} \AND \bauthor{\bsnm{Cohen},~\bfnm{Neal~J.}\binits{N.~J.}}
(\byear{1989}).
\btitle{Catastrophic interference in connectionist networks: The sequential learning problem}.
\bseries{Psychology of Learning and Motivation}.
\end{bincollection}
\endbibitem

\bibitem{molstad2022}
\begin{bmisc}[author]
\bauthor{\bsnm{Molstad},~\bfnm{Aaron~J.}\binits{A.~J.}}, \bauthor{\bsnm{Ekvall},~\bfnm{Karl~Oskar}\binits{K.~O.}} \AND \bauthor{\bsnm{Suder},~\bfnm{Piotr~M.}\binits{P.~M.}}
(\byear{2022}).
\btitle{Direct covariance matrix estimation with compositional data. \textit{arXiv preprint \href{https://arxiv.org/pdf/2212.09833.pdf}{arXiv 2212.09833}}}.
\end{bmisc}
\endbibitem

\bibitem{moran2022identifiable}
\begin{barticle}[author]
\bauthor{\bsnm{Moran},~\bfnm{Gemma~Elyse}\binits{G.~E.}}, \bauthor{\bsnm{Sridhar},~\bfnm{Dhanya}\binits{D.}}, \bauthor{\bsnm{Wang},~\bfnm{Yixin}\binits{Y.}} \AND \bauthor{\bsnm{Blei},~\bfnm{David}\binits{D.}}
(\byear{2022}).
\btitle{Identifiable deep generative models via sparse decoding}.
\bjournal{Transactions on Machine Learning Research}.
\end{barticle}
\endbibitem

\bibitem{Muller_2004}
\begin{barticle}[author]
\bauthor{\bsnm{Müller},~\bfnm{Peter}\binits{P.}}, \bauthor{\bsnm{Quintana},~\bfnm{Fernando}\binits{F.}} \AND \bauthor{\bsnm{Rosner},~\bfnm{Gary}\binits{G.}}
(\byear{2004}).
\btitle{A method for combining inference across related nonparametric Bayesian models}.
\bjournal{Journal of the Royal Statistical Society: Series B (Statistical Methodology)}.
\end{barticle}
\endbibitem

\bibitem{survey_tl_2020}
\begin{barticle}[author]
\bauthor{\bsnm{Niu},~\bfnm{Shuteng}\binits{S.}}, \bauthor{\bsnm{Liu},~\bfnm{Yongxin}\binits{Y.}}, \bauthor{\bsnm{Wang},~\bfnm{Jian}\binits{J.}} \AND \bauthor{\bsnm{Song},~\bfnm{Houbing}\binits{H.}}
(\byear{2020}).
\btitle{A decade survey of transfer learning (2010–2020)}.
\bjournal{IEEE Transactions on Artificial Intelligence}.
\end{barticle}
\endbibitem

\bibitem{survey_transfer_learning}
\begin{barticle}[author]
\bauthor{\bsnm{Pan},~\bfnm{Sinno~Jialin}\binits{S.~J.}} \AND \bauthor{\bsnm{Yang},~\bfnm{Qiang}\binits{Q.}}
(\byear{2010}).
\btitle{A survey on transfer learning}.
\bjournal{IEEE Transactions on Knowledge and Data Engineering}.
\end{barticle}
\endbibitem

\bibitem{amortized_meta_few_shot}
\begin{barticle}[author]
\bauthor{\bsnm{Patacchiola},~\bfnm{Massimiliano}\binits{M.}}, \bauthor{\bsnm{Turner},~\bfnm{Jack}\binits{J.}}, \bauthor{\bsnm{Crowley},~\bfnm{Elliot~J.}\binits{E.~J.}}, \bauthor{\bsnm{O\textquotesingle~Boyle},~\bfnm{Michael}\binits{M.}} \AND \bauthor{\bsnm{Storkey},~\bfnm{Amos~J}\binits{A.~J.}}
(\byear{2020}).
\btitle{Bayesian meta-learning for the few-shot setting via deep kernels}.
\bjournal{Advances in Neural Information Processing Systems}.
\end{barticle}
\endbibitem

\bibitem{blessing_dim}
\begin{barticle}[author]
\bauthor{\bsnm{Quefeng~Li},~\bfnm{Jianqing~Fan}\binits{J.~F.} \bsuffix{Guang~Cheng}} \AND \bauthor{\bsnm{Wang},~\bfnm{Yuyan}\binits{Y.}}
(\byear{2018}).
\btitle{Embracing the blessing of dimensionality in factor models}.
\bjournal{Journal of the American Statistical Association}.
\end{barticle}
\endbibitem

\bibitem{ravi2018amortized}
\begin{barticle}[author]
\bauthor{\bsnm{Ravi},~\bfnm{Sachin}\binits{S.}} \AND \bauthor{\bsnm{Beatson},~\bfnm{Alex}\binits{A.}}
(\byear{2019}).
\btitle{Amortized Bayesian meta-learning}.
\bjournal{International Conference on Learning Representations}.
\end{barticle}
\endbibitem

\bibitem{trans_clime}
\begin{barticle}[author]
\bauthor{\bsnm{Sai~Li},~\bfnm{T.~Tony~Cai}\binits{T.~T.~C.}} \AND \bauthor{\bsnm{Li},~\bfnm{Hongzhe}\binits{H.}}
(\byear{2022}).
\btitle{Transfer learning in large-scale Gaussian graphical models with false discovery rate control}.
\bjournal{Journal of the American Statistical Association}.
\end{barticle}
\endbibitem

\bibitem{hier_bayes_one_shot}
\begin{barticle}[author]
\bauthor{\bsnm{Salakhutdinov},~\bfnm{Ruslan}\binits{R.}}, \bauthor{\bsnm{Tenenbaum},~\bfnm{Joshua}\binits{J.}} \AND \bauthor{\bsnm{Torralba},~\bfnm{Antonio}\binits{A.}}
(\byear{2012}).
\btitle{One-shot learning with a hierarchical nonparametric Bayesian model}.
\bjournal{Proceedings of ICML Workshop on Unsupervised and Transfer Learning}.
\end{barticle}
\endbibitem

\bibitem{pan_cancer_surv2}
\begin{barticle}[author]
\bauthor{\bsnm{Samorodnitsky},~\bfnm{Sarah}\binits{S.}}, \bauthor{\bsnm{Hoadley},~\bfnm{Katherine}\binits{K.}} \AND \bauthor{\bsnm{Lock},~\bfnm{Eric}\binits{E.}}
(\byear{2020}).
\btitle{A pan-cancer and polygenic Bayesian hierarchical model for the effect of somatic mutations on survival}.
\bjournal{Cancer Informatics}.
\end{barticle}
\endbibitem

\bibitem{deep_tl_imaging}
\begin{barticle}[author]
\bauthor{\bsnm{Shin},~\bfnm{Hoo-Chang}\binits{H.-C.}}, \bauthor{\bsnm{Roth},~\bfnm{Holger~R}\binits{H.~R.}}, \bauthor{\bsnm{Gao},~\bfnm{Mingchen}\binits{M.}}, \bauthor{\bsnm{Lu},~\bfnm{Le}\binits{L.}}, \bauthor{\bsnm{Xu},~\bfnm{Ziyue}\binits{Z.}}, \bauthor{\bsnm{Nogues},~\bfnm{Isabella}\binits{I.}}, \bauthor{\bsnm{Yao},~\bfnm{Jianhua}\binits{J.}}, \bauthor{\bsnm{Mollura},~\bfnm{Daniel}\binits{D.}} \AND \bauthor{\bsnm{Summers},~\bfnm{Ronald~M}\binits{R.~M.}}
\btitle{Deep convolutional neural networks for {computer-aided} detection: {CNN} architectures, dataset characteristics and transfer learning}.
\bjournal{IEEE Transactions on Medical Imaging}.
\end{barticle}
\endbibitem

\bibitem{pretrain_loss_deep}
\begin{barticle}[author]
\bauthor{\bsnm{Shwartz-Ziv},~\bfnm{Ravid}\binits{R.}}, \bauthor{\bsnm{Goldblum},~\bfnm{Micah}\binits{M.}}, \bauthor{\bsnm{Souri},~\bfnm{Hossein}\binits{H.}}, \bauthor{\bsnm{Kapoor},~\bfnm{Sanyam}\binits{S.}}, \bauthor{\bsnm{Zhu},~\bfnm{Chen}\binits{C.}}, \bauthor{\bsnm{LeCun},~\bfnm{Yann}\binits{Y.}} \AND \bauthor{\bsnm{Wilson},~\bfnm{Andrew~G}\binits{A.~G.}}
(\byear{2022}).
\btitle{Pre-train your loss: Easy Bayesian transfer learning with informative priors}.
\bjournal{Advances in Neural Information Processing Systems}.
\end{barticle}
\endbibitem

\bibitem{pac_bayes_multiclass}
\begin{barticle}[author]
\bauthor{\bsnm{Sicilia},~\bfnm{Anthony}\binits{A.}}, \bauthor{\bsnm{Atwell},~\bfnm{Katherine}\binits{K.}}, \bauthor{\bsnm{Alikhani},~\bfnm{Malihe}\binits{M.}} \AND \bauthor{\bsnm{Hwang},~\bfnm{Seong~Jae}\binits{S.~J.}}
(\byear{2022}).
\btitle{\uppercase{PAC-B}ayesian domain adaptation bounds for multiclass learners}.
\bjournal{Uncertainty in Artificial Intelligence}.
\end{barticle}
\endbibitem

\bibitem{song2019harmonized}
\begin{barticle}[author]
\bauthor{\bsnm{Song},~\bfnm{Guoli}\binits{G.}}, \bauthor{\bsnm{Wang},~\bfnm{Shuhui}\binits{S.}}, \bauthor{\bsnm{Huang},~\bfnm{Qingming}\binits{Q.}} \AND \bauthor{\bsnm{Tian},~\bfnm{Qi}\binits{Q.}}
(\byear{2019}).
\btitle{Harmonized multimodal learning with Gaussian process latent variable models}.
\bjournal{IEEE Transactions on Pattern Analysis and Machine Intelligence}.
\end{barticle}
\endbibitem

\bibitem{Sotiropoulos2016-jj}
\begin{barticle}[author]
\bauthor{\bsnm{Sotiropoulos},~\bfnm{Stamatios~N}\binits{S.~N.}}, \bauthor{\bsnm{Hern{\'a}ndez-Fern{\'a}ndez},~\bfnm{Mois{\'e}s}\binits{M.}}, \bauthor{\bsnm{Vu},~\bfnm{An~T}\binits{A.~T.}}, \bauthor{\bsnm{Andersson},~\bfnm{Jesper~L}\binits{J.~L.}}, \bauthor{\bsnm{Moeller},~\bfnm{Steen}\binits{S.}}, \bauthor{\bsnm{Yacoub},~\bfnm{Essa}\binits{E.}}, \bauthor{\bsnm{Lenglet},~\bfnm{Christophe}\binits{C.}}, \bauthor{\bsnm{Ugurbil},~\bfnm{Kamil}\binits{K.}}, \bauthor{\bsnm{Behrens},~\bfnm{Timothy E~J}\binits{T.~E.~J.}} \AND \bauthor{\bsnm{Jbabdi},~\bfnm{Saad}\binits{S.}}
(\byear{2016}).
\btitle{Fusion in diffusion {MRI} for improved fibre orientation estimation: An application to the {3T} and {7T} data of the Human Connectome Project}.
\bjournal{Neuroimage}.
\end{barticle}
\endbibitem

\bibitem{Sotiropoulos2013-cb}
\begin{barticle}[author]
\bauthor{\bsnm{Sotiropoulos},~\bfnm{Stamatios~N}\binits{S.~N.}}, \bauthor{\bsnm{Jbabdi},~\bfnm{Saad}\binits{S.}}, \bauthor{\bsnm{Andersson},~\bfnm{Jesper~L}\binits{J.~L.}}, \bauthor{\bsnm{Woolrich},~\bfnm{Mark~W}\binits{M.~W.}}, \bauthor{\bsnm{Ugurbil},~\bfnm{Kamil}\binits{K.}} \AND \bauthor{\bsnm{Behrens},~\bfnm{Timothy E~J}\binits{T.~E.~J.}}
(\byear{2013}).
\btitle{{RubiX}: combining spatial resolutions for Bayesian inference of crossing fibers in diffusion {MRI}}.
\bjournal{IEEE Transactions on Medical Imaging}.
\end{barticle}
\endbibitem

\bibitem{a0_criterion_2}
\begin{barticle}[author]
\bauthor{\bsnm{Spiegelhalter},~\bfnm{David~J.}\binits{D.~J.}}, \bauthor{\bsnm{Best},~\bfnm{Nicola~G.}\binits{N.~G.}}, \bauthor{\bsnm{Carlin},~\bfnm{Bradley~P.}\binits{B.~P.}} \AND \bauthor{\bsnm{Van Der~Linde},~\bfnm{Angelika}\binits{A.}}
(\byear{2002}).
\btitle{Bayesian measures of model complexity and fit}.
\bjournal{Journal of the Royal Statistical Society: Series B (Statistical Methodology)}.
\end{barticle}
\endbibitem

\bibitem{deep_tl_survey}
\begin{barticle}[author]
\bauthor{\bsnm{Tan},~\bfnm{Chuanqi}\binits{C.}}, \bauthor{\bsnm{Sun},~\bfnm{Fuchun}\binits{F.}}, \bauthor{\bsnm{Kong},~\bfnm{Tao}\binits{T.}}, \bauthor{\bsnm{Zhang},~\bfnm{Wenchang}\binits{W.}}, \bauthor{\bsnm{Yang},~\bfnm{Chao}\binits{C.}} \AND \bauthor{\bsnm{Liu},~\bfnm{Chunfang}\binits{C.}}
(\byear{2018}).
\btitle{A survey on deep transfer learning}.
\bjournal{Artificial Neural Networks and Machine Learning}.
\end{barticle}
\endbibitem

\bibitem{prot_network}
\begin{barticle}[author]
\bauthor{\bsnm{Tan},~\bfnm{Linda S.~L.}\binits{L.~S.~L.}}, \bauthor{\bsnm{Jasra},~\bfnm{Ajay}\binits{A.}}, \bauthor{\bsnm{Iorio},~\bfnm{Maria~De}\binits{M.~D.}} \AND \bauthor{\bsnm{Ebbels},~\bfnm{Timothy M.~D.}\binits{T.~M.~D.}}
(\byear{2017}).
\btitle{{Bayesian inference for multiple Gaussian graphical models with application to metabolic association networks}}.
\bjournal{The Annals of Applied Statistics}.
\end{barticle}
\endbibitem

\bibitem{Teh_2006}
\begin{barticle}[author]
\bauthor{\bsnm{Teh},~\bfnm{Yee~Whye}\binits{Y.~W.}}, \bauthor{\bsnm{Jordan},~\bfnm{Michael~I.}\binits{M.~I.}}, \bauthor{\bsnm{Beal},~\bfnm{Matthew~J.}\binits{M.~J.}} \AND \bauthor{\bsnm{Blei},~\bfnm{David~M.}\binits{D.~M.}}
(\byear{2006}).
\btitle{Hierarchical Dirichlet processes}.
\bjournal{Journal of the American Statistical Association}.
\end{barticle}
\endbibitem

\bibitem{clime}
\begin{barticle}[author]
\bauthor{\bsnm{Tony~Cai},~\bfnm{Weidong~Liu}\binits{W.~L.}} \AND \bauthor{\bsnm{Luo},~\bfnm{Xi}\binits{X.}}
(\byear{2011}).
\btitle{A constrained $L_1$ minimization approach to sparse precision matrix estimation}.
\bjournal{Journal of the American Statistical Association}.
\end{barticle}
\endbibitem

\bibitem{deep_tl_tzeng}
\begin{barticle}[author]
\bauthor{\bsnm{Tzeng},~\bfnm{Eric}\binits{E.}}, \bauthor{\bsnm{Hoffman},~\bfnm{Judy}\binits{J.}}, \bauthor{\bsnm{Darrell},~\bfnm{Trevor}\binits{T.}} \AND \bauthor{\bsnm{Saenko},~\bfnm{Kate}\binits{K.}}
(\byear{2015}).
\btitle{Simultaneous Deep Transfer Across Domains and Tasks}.
\bjournal{IEEE International Conference on Computer Vision}.
\end{barticle}
\endbibitem

\bibitem{Vanschoren2019}
\begin{binbook}[author]
\bauthor{\bsnm{Vanschoren},~\bfnm{Joaquin}\binits{J.}}
(\byear{2019}).
\btitle{Meta-Learning}
In \bbooktitle{Automated Machine Learning: Methods, Systems, Challenges}.
\end{binbook}
\endbibitem

\bibitem{de_vito_2021}
\begin{barticle}[author]
\bauthor{\bsnm{Vito},~\bfnm{Roberta~De}\binits{R.~D.}}, \bauthor{\bsnm{Bellio},~\bfnm{Ruggero}\binits{R.}}, \bauthor{\bsnm{Trippa},~\bfnm{Lorenzo}\binits{L.}} \AND \bauthor{\bsnm{Parmigiani},~\bfnm{Giovanni}\binits{G.}}
(\byear{2021}).
\btitle{{Bayesian multistudy factor analysis for high-throughput biological data}}.
\bjournal{The Annals of Applied Statistics}.
\end{barticle}
\endbibitem

\bibitem{Wang_Pineau_2015}
\begin{barticle}[author]
\bauthor{\bsnm{Wang},~\bfnm{Boyu}\binits{B.}} \AND \bauthor{\bsnm{Pineau},~\bfnm{Joelle}\binits{J.}}
(\byear{2015}).
\btitle{Online boosting algorithms for anytime transfer and multitask learning}.
\bjournal{AAAI Conference on Artificial Intelligence}.
\end{barticle}
\endbibitem

\bibitem{posterior_collapse}
\begin{barticle}[author]
\bauthor{\bsnm{Wang},~\bfnm{Yixin}\binits{Y.}}, \bauthor{\bsnm{Blei},~\bfnm{David}\binits{D.}} \AND \bauthor{\bsnm{Cunningham},~\bfnm{John~P}\binits{J.~P.}}
(\byear{2021}).
\btitle{Posterior collapse and latent variable non-identifiability}.
\bjournal{Advances in Neural Information Processing Systems}.
\end{barticle}
\endbibitem

\bibitem{post_collapse_2}
\begin{barticle}[author]
\bauthor{\bsnm{Wang},~\bfnm{Zihao}\binits{Z.}} \AND \bauthor{\bsnm{Ziyin},~\bfnm{Liu}\binits{L.}}
(\byear{2022}).
\btitle{Posterior collapse of a linear latent variable model}.
\bjournal{Advances in Neural Information Processing Systems}.
\end{barticle}
\endbibitem

\bibitem{multitask_gp}
\begin{barticle}[author]
\bauthor{\bsnm{Wilson},~\bfnm{Andrew~Gordon}\binits{A.~G.}}, \bauthor{\bsnm{Knowles},~\bfnm{David~A.}\binits{D.~A.}} \AND \bauthor{\bsnm{Ghahramani},~\bfnm{Zoubin}\binits{Z.}}
(\byear{2012}).
\btitle{Gaussian process regression networks}.
\bjournal{International Conference on Machine Learning}.
\end{barticle}
\endbibitem

\bibitem{bayes_language_da}
\begin{barticle}[author]
\bauthor{\bsnm{Wood},~\bfnm{Frank}\binits{F.}} \AND \bauthor{\bsnm{Teh},~\bfnm{Yee~Whye}\binits{Y.~W.}}
(\byear{2009}).
\btitle{A hierarchical nonparametric Bayesian approach to statistical language model domain adaptation}.
\bjournal{International Conference on Artificial Intelligence and Statistics}.
\end{barticle}
\endbibitem

\bibitem{xu2022proximal}
\begin{barticle}[author]
\bauthor{\bsnm{Xu},~\bfnm{Jason}\binits{J.}} \AND \bauthor{\bsnm{Lange},~\bfnm{Kenneth}\binits{K.}}
(\byear{2022}).
\btitle{A proximal distance algorithm for likelihood-based sparse covariance estimation}.
\bjournal{Biometrika}.
\end{barticle}
\endbibitem

\bibitem{reinforced_cont_learning}
\begin{barticle}[author]
\bauthor{\bsnm{Xu},~\bfnm{Ju}\binits{J.}} \AND \bauthor{\bsnm{Zhu},~\bfnm{Zhanxing}\binits{Z.}}
(\byear{2018}).
\btitle{Reinforced continual learning}.
\bjournal{Advances in Neural Information Processing Systems}.
\end{barticle}
\endbibitem

\bibitem{xu2023identifiable}
\begin{bmisc}[author]
\bauthor{\bsnm{Xu},~\bfnm{Maoran}\binits{M.}}, \bauthor{\bsnm{Herring},~\bfnm{Amy~H.}\binits{A.~H.}} \AND \bauthor{\bsnm{Dunson},~\bfnm{David~B.}\binits{D.~B.}}
(\byear{2023}).
\btitle{Identifiable and interpretable nonparametric factor analysis. \textit{arXiv preprint \href{https://arxiv.org/pdf/2311.08254.pdf}{arXiv 2311.08254}}}.
\end{bmisc}
\endbibitem

\bibitem{xuan2021bayesian}
\begin{barticle}[author]
\bauthor{\bsnm{Xuan},~\bfnm{Junyu}\binits{J.}}, \bauthor{\bsnm{Lu},~\bfnm{Jie}\binits{J.}} \AND \bauthor{\bsnm{Zhang},~\bfnm{Guangquan}\binits{G.}}
(\byear{2021}).
\btitle{Bayesian transfer learning: An overview of probabilistic graphical models for transfer learning. \textit{arXiv preprint \href{https://arxiv.org/pdf/2109.13233.pdf}{arXiv 2109.13233}}}.
\end{barticle}
\endbibitem

\bibitem{yang_zhang_dai_pan_2020}
\begin{bbook}[author]
\bauthor{\bsnm{Yang},~\bfnm{Qiang}\binits{Q.}}, \bauthor{\bsnm{Zhang},~\bfnm{Yu}\binits{Y.}}, \bauthor{\bsnm{Dai},~\bfnm{Wenyuan}\binits{W.}} \AND \bauthor{\bsnm{Pan},~\bfnm{Sinno~Jialin}\binits{S.~J.}}
(\byear{2020}).
\btitle{Transfer learning}.
\bpublisher{Cambridge University Press}.
\end{bbook}
\endbibitem

\bibitem{bayesian_metalearning}
\begin{barticle}[author]
\bauthor{\bsnm{Yoon},~\bfnm{Jaesik}\binits{J.}}, \bauthor{\bsnm{Kim},~\bfnm{Taesup}\binits{T.}}, \bauthor{\bsnm{Dia},~\bfnm{Ousmane}\binits{O.}}, \bauthor{\bsnm{Kim},~\bfnm{Sungwoong}\binits{S.}}, \bauthor{\bsnm{Bengio},~\bfnm{Yoshua}\binits{Y.}} \AND \bauthor{\bsnm{Ahn},~\bfnm{Sungjin}\binits{S.}}
(\byear{2018}).
\btitle{Bayesian model-agnostic meta-learning}.
\bjournal{Advances in Neural Information Processing Systems}.
\end{barticle}
\endbibitem

\bibitem{deep_tl_1}
\begin{barticle}[author]
\bauthor{\bsnm{Yosinski},~\bfnm{Jason}\binits{J.}}, \bauthor{\bsnm{Clune},~\bfnm{Jeff}\binits{J.}}, \bauthor{\bsnm{Bengio},~\bfnm{Yoshua}\binits{Y.}} \AND \bauthor{\bsnm{Lipson},~\bfnm{Hod}\binits{H.}}
(\byear{2014}).
\btitle{How transferable are features in deep neural networks?}
\bjournal{Advances in Neural Information Processing Systems}.
\end{barticle}
\endbibitem

\bibitem{aggregated_data_gp}
\begin{barticle}[author]
\bauthor{\bsnm{Yousefi},~\bfnm{Fariba}\binits{F.}}, \bauthor{\bsnm{Smith},~\bfnm{Michael~T}\binits{M.~T.}} \AND \bauthor{\bsnm{\'{A}lvarez},~\bfnm{Mauricio}\binits{M.}}
(\byear{2019}).
\btitle{Multi-task learning for aggregated data using Gaussian processes}.
\bjournal{Advances in Neural Information Processing Systems}.
\end{barticle}
\endbibitem

\bibitem{cao2019}
\begin{barticle}[author]
\bauthor{\bsnm{Yuanpei~Cao},~\bfnm{Wei~Lin}\binits{W.~L.}} \AND \bauthor{\bsnm{Li},~\bfnm{Hongzhe}\binits{H.}}
(\byear{2019}).
\btitle{Large covariance estimation for compositional data via composition-adjusted thresholding}.
\bjournal{Journal of the American Statistical Association}.
\end{barticle}
\endbibitem

\bibitem{continual_metalearning}
\begin{barticle}[author]
\bauthor{\bsnm{Zhang},~\bfnm{Qiang}\binits{Q.}}, \bauthor{\bsnm{Fang},~\bfnm{Jinyuan}\binits{J.}}, \bauthor{\bsnm{Meng},~\bfnm{Zaiqiao}\binits{Z.}}, \bauthor{\bsnm{Liang},~\bfnm{Shangsong}\binits{S.}} \AND \bauthor{\bsnm{Yilmaz},~\bfnm{Emine}\binits{E.}}
(\byear{2021}).
\btitle{Variational continual Bayesian meta-learning}.
\bjournal{Advances in Neural Information Processing Systems}.
\end{barticle}
\endbibitem

\bibitem{negative_transfer_survey}
\begin{barticle}[author]
\bauthor{\bsnm{Zhang},~\bfnm{Wen}\binits{W.}}, \bauthor{\bsnm{Deng},~\bfnm{Lingfei}\binits{L.}}, \bauthor{\bsnm{Zhang},~\bfnm{Lei}\binits{L.}} \AND \bauthor{\bsnm{Wu},~\bfnm{Dongrui}\binits{D.}}
(\byear{2023}).
\btitle{A survey on negative transfer}.
\bjournal{IEEE/CAA Journal of Automatica Sinica}.
\end{barticle}
\endbibitem

\bibitem{continual_learning_2}
\begin{barticle}[author]
\bauthor{\bsnm{Zhao},~\bfnm{Tingting}\binits{T.}}, \bauthor{\bsnm{Wang},~\bfnm{Zifeng}\binits{Z.}}, \bauthor{\bsnm{Masoomi},~\bfnm{Aria}\binits{A.}} \AND \bauthor{\bsnm{Dy},~\bfnm{Jennifer}\binits{J.}}
(\byear{2022}).
\btitle{Deep Bayesian unsupervised lifelong learning}.
\bjournal{Neural Networks}.
\end{barticle}
\endbibitem

\bibitem{bayes_cov_shift}
\begin{barticle}[author]
\bauthor{\bsnm{Zhou},~\bfnm{Aurick}\binits{A.}} \AND \bauthor{\bsnm{Levine},~\bfnm{Sergey}\binits{S.}}
(\byear{2021}).
\btitle{Bayesian adaptation for covariate shift}.
\bjournal{Advances in Neural Information Processing Systems}.
\end{barticle}
\endbibitem

\bibitem{multiple_learners_graph}
\begin{barticle}[author]
\bauthor{\bsnm{Zhou},~\bfnm{Jiaying}\binits{J.}}, \bauthor{\bsnm{Ding},~\bfnm{Jie}\binits{J.}}, \bauthor{\bsnm{Tan},~\bfnm{Kean~Ming}\binits{K.~M.}} \AND \bauthor{\bsnm{Tarokh},~\bfnm{Vahid}\binits{V.}}
(\byear{2021}).
\btitle{Model linkage selection for cooperative learning}.
\bjournal{Journal of Machine Learning Research}.
\end{barticle}
\endbibitem

\bibitem{ieee_survey_tl}
\begin{barticle}[author]
\bauthor{\bsnm{Zhuang},~\bfnm{Fuzhen}\binits{F.}}, \bauthor{\bsnm{Qi},~\bfnm{Zhiyuan}\binits{Z.}}, \bauthor{\bsnm{Duan},~\bfnm{Keyu}\binits{K.}}, \bauthor{\bsnm{Xi},~\bfnm{Dongbo}\binits{D.}}, \bauthor{\bsnm{Zhu},~\bfnm{Yongchun}\binits{Y.}}, \bauthor{\bsnm{Zhu},~\bfnm{Hengshu}\binits{H.}}, \bauthor{\bsnm{Xiong},~\bfnm{Hui}\binits{H.}} \AND \bauthor{\bsnm{He},~\bfnm{Qing}\binits{Q.}}
(\byear{2021}).
\btitle{A comprehensive survey on transfer learning}.
\bjournal{Proceedings of the IEEE}.
\end{barticle}
\endbibitem

\end{thebibliography}

\end{document}